\documentclass[10pt,twocolumn,letterpaper]{article}

\usepackage{cvpr}
\usepackage{times}
\usepackage{epsfig}
\usepackage{graphicx}
\usepackage{amsmath}
\usepackage{amssymb}
\usepackage{cite}
\usepackage{bm}
\usepackage{microtype}
\usepackage[tight,footnotesize]{subfigure}
\usepackage{multirow}
\usepackage{floatrow}
\usepackage{color}
\usepackage{colortbl}
\usepackage{array}
\usepackage{caption}
\usepackage{url}
\definecolor{mygray}{gray}{.9}

\usepackage[breaklinks=true,bookmarks=false]{hyperref}

\cvprfinalcopy 

\def\cvprPaperID{****} 

\ifcvprfinal\fi
\begin{document}

\title{Person Re-identification in the Wild}

\author{Liang Zheng$^1$, Hengheng Zhang$^2$, Shaoyan Sun$^3$, Manmohan Chandraker$^4$, Yi Yang$^1$, Qi Tian$^2$ \\
 $^1$University of Technology Sydney \qquad $^2$UTSA   \qquad $^3$USTC \qquad $^4$UCSD\\
{\tt\small {\{liangzheng06,manu.chandraker,yee.i.yang,wywqtian\}}@gmail.com }
 }


\maketitle
\thispagestyle{empty}
\begin{abstract}
 This paper\footnote{\noindent{\scriptsize L. Zheng, H. Zhang and S. Sun contribute equally.
 This work was partially supported by the Google Faculty Award and the Data to Decisions Cooperative Research Centre. This work was supported in part to Dr. Qi Tian by ARO grant W911NF-15-1-0290 and Faculty Research Gift Awards by NEC Laboratories of America and Blippar. This work was supported in part by National Science Foundation of China (NSFC) 61429201. Project page: \url{http://www.liangzheng.com.cn}}} presents a novel large-scale dataset and comprehensive baselines for end-to-end pedestrian detection and person recognition in raw video frames. Our baselines address three issues: the performance of various combinations of detectors and recognizers, mechanisms for pedestrian detection to help improve overall re-identification (re-ID) accuracy and assessing the effectiveness of different detectors for re-ID. We make three distinct contributions. First, a new dataset, PRW, is introduced to evaluate \textbf{P}erson \textbf{R}e-identification in the \textbf{W}ild, using videos acquired through six near-synchronized cameras. It contains 932 identities and 11,816 frames in which pedestrians are annotated with their bounding box positions and identities. Extensive benchmarking results are presented on this dataset. Second, we show that pedestrian detection aids re-ID through two simple yet effective improvements: a cascaded fine-tuning strategy that trains a detection model first and then the classification model, and a Confidence Weighted Similarity (CWS) metric that incorporates detection scores into similarity measurement. Third, we derive insights in evaluating detector performance for the particular scenario of accurate person re-ID.
\end{abstract}

\section{Introduction}
Automated entry and retail systems at theme parks, passenger flow monitoring at airports, behavior analysis for automated driving and surveillance are a few applications where detection and recognition of persons across a camera network can provide critical insights. Yet, these two problems have generally been studied in isolation within computer vision. Person re-identification (re-ID) aims to find occurrences of a query person ID in a video sequence, where state-of-the-art datasets and methods start from pre-defined bounding boxes, either hand-drawn \cite{liao2015person, ma2014covariance, xiong2014person} or automatically detected \cite{zheng2015scalable, li2014deepreid}. On the other hand, several pedestrian detectors achieve remarkable performance on benchmark datasets \cite{felzenszwalb2010object,ren2015faster}, but little analysis is available on how they can be used for person re-ID.

\begin{figure}[t]
  \centering
  \includegraphics[width=3.2in]{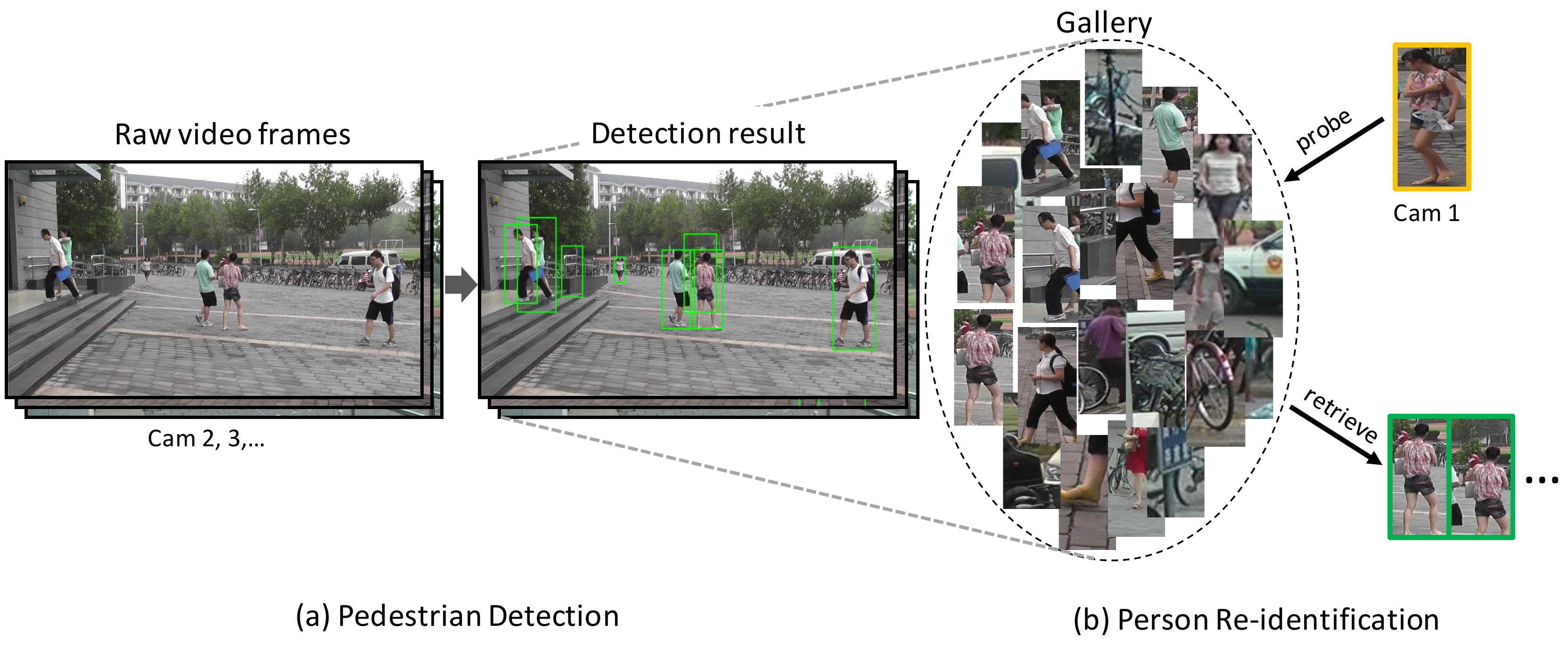}\\
  \caption{Pipeline of an end-to-end person re-ID system. It consists of two modules: pedestrian detection and person recognition (to differentiate from the overall re-ID). This paper not only benchmarks both components, but also provides novel insights in their interactions. }\label{fig:pipeline}
\end{figure}

In this paper, we propose a dataset and baselines for practical person re-ID in the wild, which moves beyond sequential application of detection and recognition. In particular, we study three aspects of the problem that have not been considered in prior works. First, we analyze the effect of the combination of various detection and recognition methods on person re-ID accuracy. Second, we study whether detection can help improve re-ID accuracy and outline methods to do so. Third, we study choices for detectors that allow for maximal gains in re-ID accuracy.

Current datasets lack annotations for such combined evaluation of person detection and re-ID. Pedestrian detection datasets, such as Caltech \cite{dollar2012pedestrian} or Inria \cite{dalal2005histograms}, typically do not have ID annotations, especially from multiple cameras. On the other hand, person re-ID datasets, such as VIPeR \cite{gray2007evaluating} or CUHK03 \cite{li2014deepreid}, usually provide just cropped bounding boxes without the complete video frames, especially at a large scale. As a consequence, a large-scale dataset that evaluates both detection and overall re-ID is needed. To address this, Section \ref{sec:dataset} presents a novel large-scale dataset called PRW that consists of $932$ identities, with bounding boxes across $11,816$ frames. The dataset comes with annotations and extensive baselines to evaluate the impacts of detection and recognition methods on person re-ID accuracy.

In Section \ref{sec:components_and_improvements}, we leverage the volume of the PRW dataset to train state-of-the-art detectors such as R-CNN \cite{girshick2014rich}, with various convolutional neural network (CNN) architectures such as AlexNet \cite{krizhevsky2012imagenet}, VGGNet \cite{simonyan2014very} and ResidualNet \cite{he2015deep}. Several well-known descriptors and distance metrics are also considered for person re-ID. However, our joint setup allows two further improvements in Section \ref{sec:improvements}. First, we propose a cascaded fine-tuning strategy to make full use of the detection data provided by PRW, which results in improved CNN embeddings. Two CNN variants, are derived \emph{w.r.t} the fine tuning strategies. Novel insights can be learned from the new fine-tuning method. Second, we propose a Confidence Weighted Similarity (CWS) metric that incorporates detection scores. Assigning lower weights to false positive detections prevents a drop in re-ID accuracy due to the increase in gallery size with the use of detectors.

Given a dataset like PRW that allows simultaneous evaluation of detection and re-ID, it is natural to consider whether any complementarity exists between the two tasks. For a particular re-ID method, it is intuitive that a better detector should yield better accuracy. But we argue that the criteria for determining a detector as better are application-dependent. Previous works in pedestrian detection \cite{dollar2012pedestrian,zhang2016far,ouyang2013joint} usually use Average Precision or Log-Average Miss Rate under IoU $> 0.5$ for evaluation. However, through extensive benchmarking on the proposed PRW dataset, we find in Section \ref{sec:experiment} that IoU $> 0.7$ is a more effective rule in indicating detector influences on re-ID accuracy. In other words, the localization ability of detectors plays a critical role in re-ID.

Figure \ref{fig:pipeline} presents the pipeline of the end-to-end re-ID system discussed in this paper. Starting from raw video frames, a gallery is created by pedestrian detectors. Given a query person-of-interest, gallery bounding boxes are ranked according to their similarity with the query. To summarize, our main contributions are:

\begin{itemize}
\item A novel large-scale dataset, Person Re-identification in the Wild (PRW), for simultaneous analysis of person detection and re-ID.
\item Comprehensive benchmarking of state-of-the-art detection and recognition methods on the PRW dataset.
\item Novel insights into how detection aids re-ID, along with an effective fine-tuning strategy and similarity measure to illustrate how they might be utilized.
\item Novel insights into the evaluation of pedestrian detectors for the specific application of person re-ID.
\end{itemize}
\section{Related Work}
\label{sec:related_work}
\vspace{-0.2cm}
{\bf An overview of existing re-ID datasets.}
In recent years, a number of person re-ID datasets have been exposed \cite{gray2007evaluating, zheng2009associating, zheng2009associating, li2013locally, li2014deepreid, zheng2015scalable,zheng2016mars}. Varying numbers of IDs and boxes exist with them (see Table \ref{table:compare_datasets}).
Despite some differences among them, a common property is that the pedestrians are confined within pre-defined bounding boxes that are either hand-drawn (\eg, VIPeR \cite{gray2007evaluating}, iLIDS \cite{zheng2009associating}, CUHK02 \cite{li2013locally}) or obtained using detectors (\eg, CUHK03 \cite{li2014deepreid}, Market-1501 \cite{zheng2015scalable} and MARS \cite{zheng2016mars}). PRW is a follow-up to our previous releases \cite{zheng2015scalable,zheng2016mars} and requires considering the entire pipeline for person re-ID from scratch.

{\bf Pedestrian detection.}
Recent pedestrian detection works  feature the ``proposal+CNN'' approach. Pedestrian detection usually employs weak pedestrian detectors as proposals, which allows achieving relatively high recall using very few proposals \cite{ouyang2012discriminative,ouyang2013modeling,ouyang2013joint,luo2014switchable}. Despite the impressive recent progress in pedestrian detection, it has been rarely considered with person re-ID as an application. This paper attempts to determine how detection can help re-ID and provide insights in assessing detector performance.

\begin{figure}[t]
  \centering
  \includegraphics[width=3.2in]{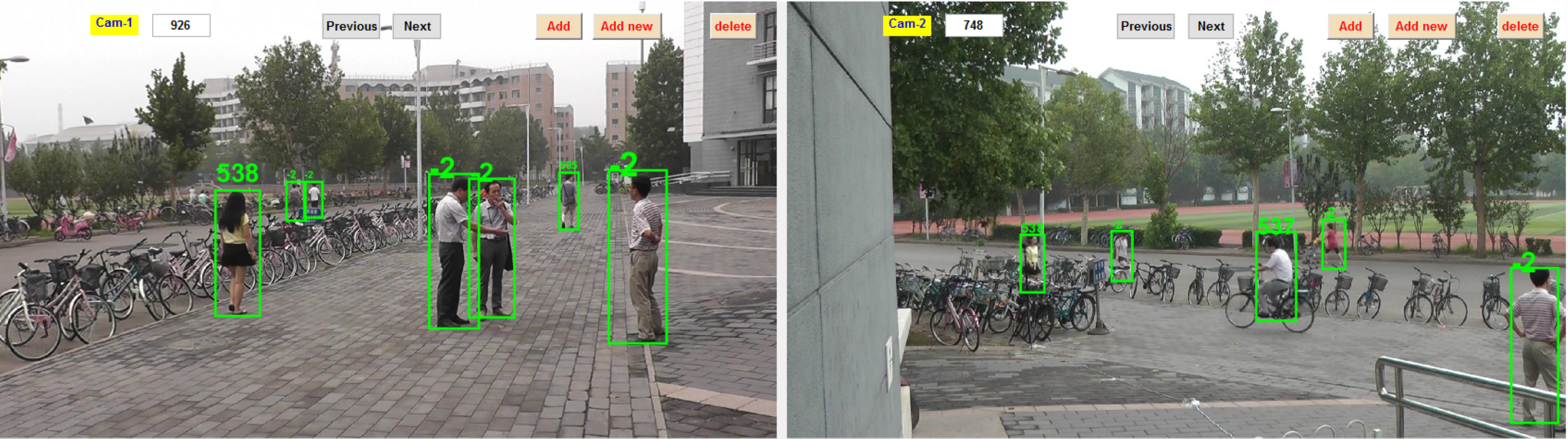}\\
  \caption{Annotation interface. All appearing pedestrians are annotated with a bounding box and ID. ID ranges from 1 to 932, and -2 stands for ambiguous persons.}\label{fig:annotation}
\end{figure}

\setlength{\tabcolsep}{6.2pt}
\begin{table*}[t]
\footnotesize
\caption{Comparing PRW with existing image-based re-ID datasets \cite{li2014deepreid, gray2007evaluating, zheng2009associating, zheng2015scalable,xu2013human,berclaz2011multiple,das2014consistent}.}
\centering
\begin{tabular}{l|ccc|ccccc}
\hline
Datasets& \textbf{PRW} &CAMPUS \cite{xu2013human} & EPFL \cite{berclaz2011multiple} &Market-1501 \cite{zheng2015scalable}&  RAiD \cite{das2014consistent} & VIPeR \cite{gray2007evaluating}&i-LIDS \cite{zheng2009associating}& CUHK03 \cite{li2014deepreid}  \\
\hline
\#frame &11,816 &214& 80 & - & -& -&-&-  \\
\#ID&932& 74& 30 & 1,501& 43 &632&119&1,360  \\
\#annotated box&34,304& 1,519& 294 & 25,259& 6,920 &1,264&476&13,164 \\
\#box per ID&36.8& 20.5 & 9.8 & 19.9&160.9 &2&2& 9.7\\
\#gallery box&100-500k& 1,519 & 294 & 19,732 &6,920& 1,264&476&13,164\\
\#camera&6& 3 & 4& 6 &4&  2 &2&2\\
\hline
\end{tabular}
\label{table:compare_datasets}
\end{table*}
{\bf Person re-ID.} Recent progress in person re-ID mainly consists in deep learning. Several  works \cite{Ahmed2015An, ding2015deep, yi2014deep,li2014deepreid, zheng2016mars} focus on learning features and metrics through the CNN framework. Formulating person re-ID as a ranking task, image pairs \cite{Ahmed2015An, yi2014deep,li2014deepreid} or triplets \cite{ding2015deep} are fed into CNN. It is also shown in \cite{zheng2016survey} that deep learning using the identification model \cite{xiao2016learning,zheng2016mars,zheng2017unlabeled} yields even higher accuracy than the siamese model. With a sufficient amount of training data per ID, we thus adopt the identification model to learn  an CNN embedding in the pedestrian subspace. We refer readers to our recent works \cite{zheng2016survey,zheng2017unlabeled} for details.

{\bf Detection and re-ID.} In our knowledge, two previous works focus on such end-to-end systems. In \cite{zhang2015beyond}, persons in photo albums are detected using poselets \cite{bourdev2009poselets} and recognition is performed using face and global signatures. However, the setting in \cite{zhang2015beyond} is not typical for person re-ID where pedestrians are observed by surveillance cameras and faces are not clear enough. In a work closer to ours, Xu \emph{et al.} \cite{xu2014person} jointly model pedestrian commonness and uniqueness, and calculate the similarity between query and each sliding window in a brute-force manner. While \cite{xu2014person} works on datasets consisting of no more than 214 video frames, it may have efficiency issues with large datasets. Departing from both works, this paper sets up a large-scale benchmark system to jointly analyze detection and re-ID performance.

Finally, we would like to refer readers to \cite{xiao2017joint}, concurrent to ours and published in the same conference, which
also releases a large dataset for end-to-end person re-ID.

\section{The PRW Dataset}
\label{sec:dataset}

\begin{figure}[t]
  \centering
  \includegraphics[width=3.2in]{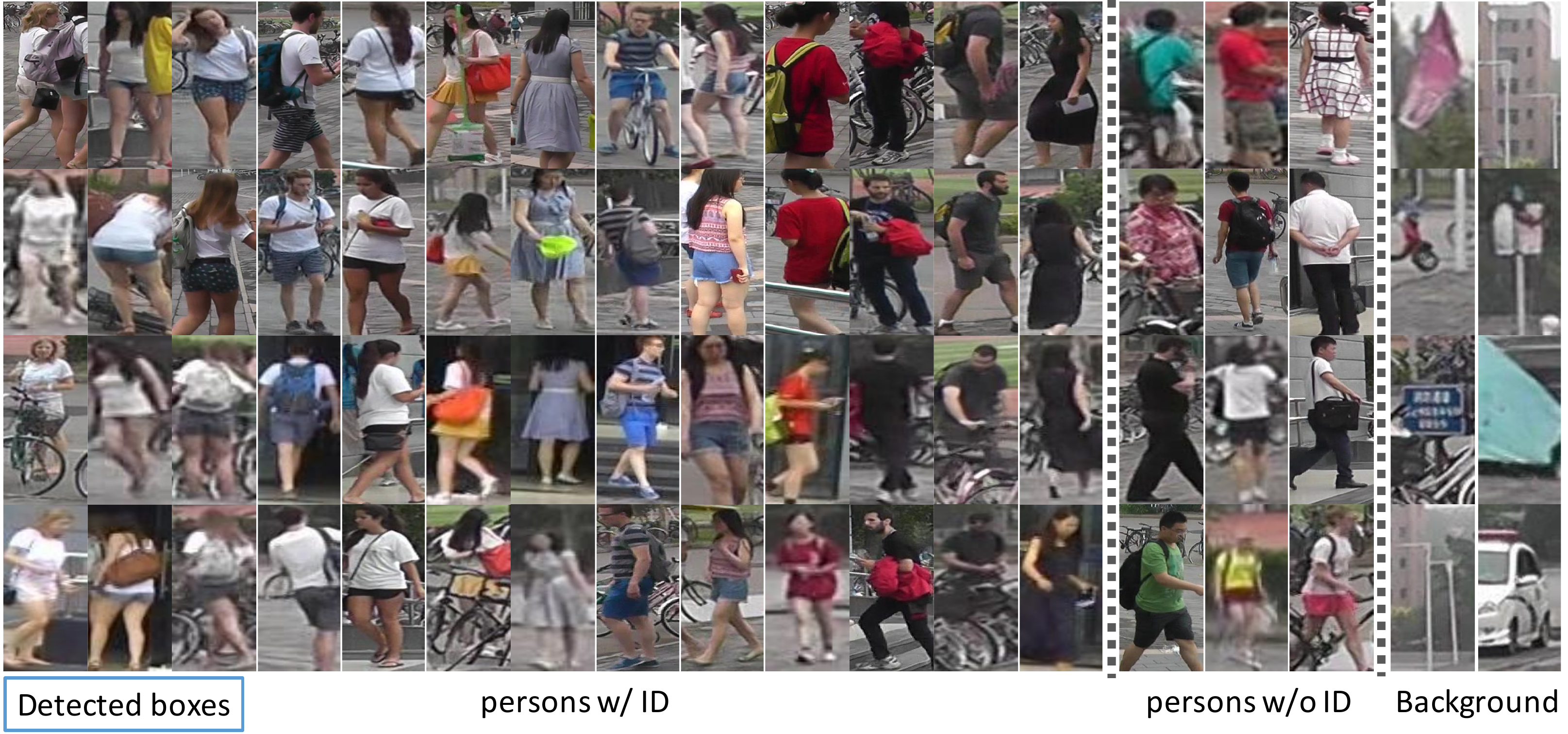}\\
  \caption{Examples of detected bounding boxes from video frames in the PRW dataset. In ``persons w$/$ID'', each column contains 4 detected boxes of the same identity from distinctive views. Column ``persons w$/$o ID'' presents persons who do not have an ID in the dataset. Column ``background'' shows false positive detection results. The detector used in this figure is DPM + RCNN (AlexNet). }\label{fig:sample_images}
\end{figure}

\subsection{Annotation Description}
The videos are collected in Tsinghua university  and are of total length 10 hours. This aims to mimic the application in which a person-of-interest goes out of the field-of-view of the current camera for a short duration and needs to be located from nearly cameras. A total of 6 cameras were used, among which 5 are 1080$\times$1920 HD and 1 is 576$\times$720 SD. The video captured by each camera is annotated every 25 frames (1 second in duration). We first manually draw a bounding box for all pedestrians who appear in the frames and then assign an ID if it exists in the Market-1501 dataset. Since all pedestrians are boxed, when we are not sure about a person's ID (ambiguity), we assign $-2$ to it. These ambiguous boxes are used in detector training and testing, but are excluded in re-ID training and testing. Figure \ref{fig:annotation} and Figure \ref{fig:sample_images} show the annotation interface and sample detected boxes, respectively.

A total of 11,816 frames are manually annotated to obtain 43,110 pedestrian bounding boxes, among which 34,304 pedestrians are annotated with an ID ranging from 1 to 932 and the rest are assigned an ID of $-2$. In Table \ref{table:compare_datasets}, we compare PRW with previous person re-ID datasets regarding numbers of frames, IDs, annotated boxes, annotated boxes per ID, gallery boxes and number of cameras. Specifically, since we densely label all the subjects, the number of boxes for each identity is almost twice that of Market-1501. Moreover, when forming the gallery, the detectors produce 100k-500k boxes depending on the threshold. The distinctive feature enabled by the PRW dataset is the end-to-end evaluation of person re-ID systems. This dataset provides the original video frames along with hand-drawn ground truth bounding boxes, which makes it feasible to evaluate both pedestrian detection and person re-ID. But more importantly, PRW enables assessing the influence of pedestrian detection on person re-ID, which is a topic of great interest for practical applications but rarely considered in previous literature.

\makeatother
\begin{figure} [t]
\centering
\subfigure[height distribution]{\label{fig:height}%
\includegraphics[width=1.6in]{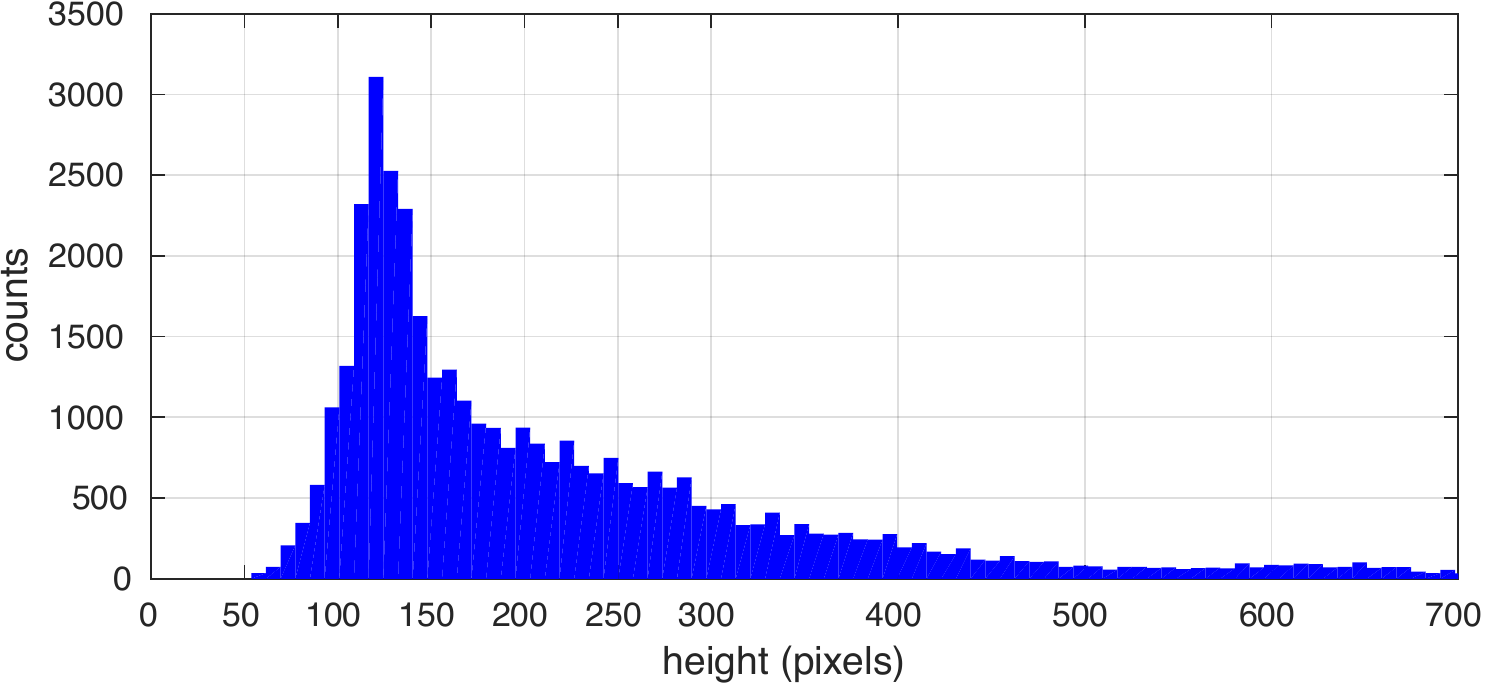}}
 \subfigure[aspect ratio distribution]{\label{fig:aspect_ratio}%
\includegraphics[width=1.6in]{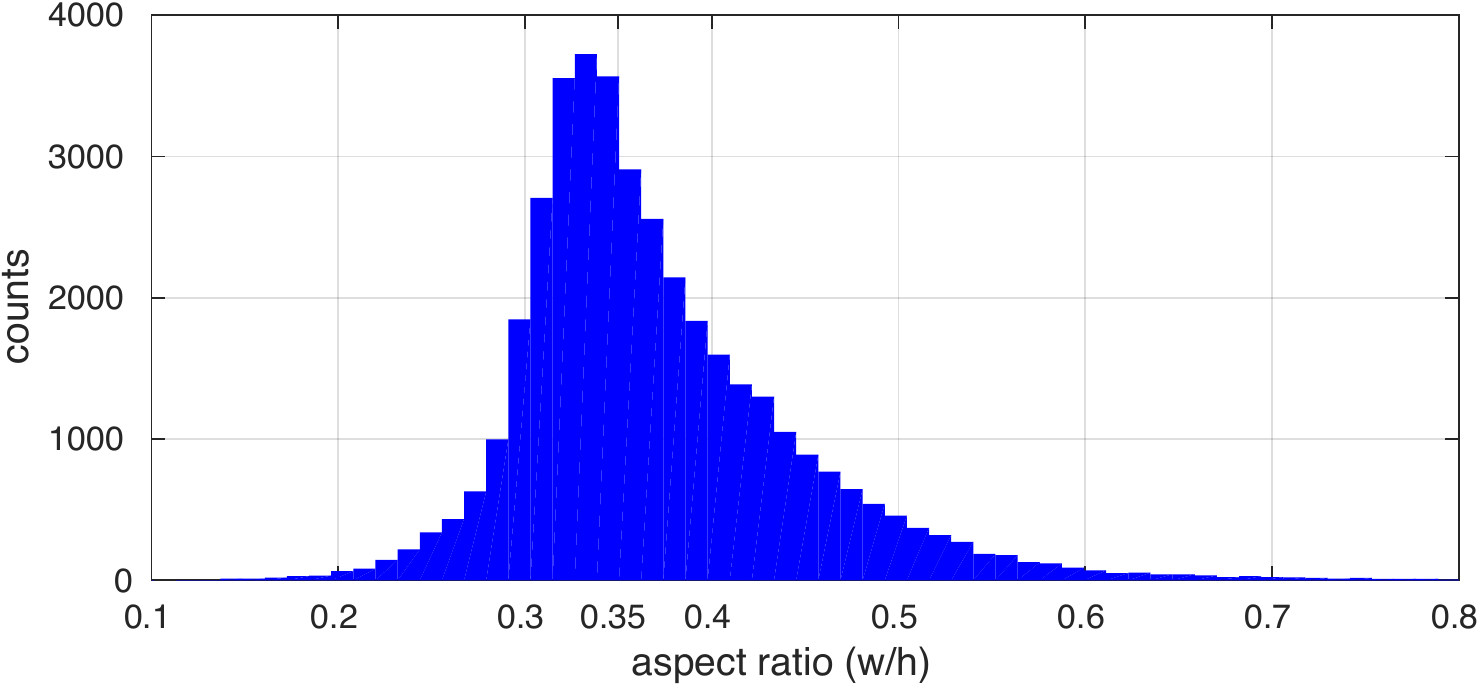}}
\caption{Distribution of pedestrian height and aspect ratio (width/height) in the PRW dataset.}
\label{fig:statistics}
\end {figure}

\subsection{Evaluation Protocols}
The PRW dataset is divided into a training set with $5,704$ frames and $482$ IDs and a test set with $6,112$ frames and $450$ IDs. We choose this split since it enables the minimum ID overlap between training and testing sets. Detailed statistics of the splits are presented in Table \ref{table:statistics_PRW}.


{\bf Pedestrian Detection.}
A number of popular pedestrian datasets exist, to name a few, INRIA \cite{dalal2005histograms}, Caltech \cite{dollar2012pedestrian} and KITTI \cite{geiger2012we}. The INRIA dataset contains 1,805 128$\times$64 pedestrian images cropped from personal photos; the Caltech dataset provides $\sim$350k bounding boxes from $\sim$132k frames; the KITTI dataset has 80k labels for the pedestrian class. With respect to the number of annotations, PRW ($\sim$43k boxes) is a medium-sized dataset for pedestrian detection. The training and testing splits are as described above and in Table \ref{table:statistics_PRW}. Following the protocols in KITTI as well as generic object detection, we mainly use the precision-recall curve and average precision to evaluate detection performance. We also report the log-average miss rate (MR) proposed in \cite{dollar2012pedestrian}. The former calculate the average precision corresponding to ten recalls uniformly sampled from $[0, 1]$ \cite{girshick2014rich}, while MR is the average miss rate at 9 False Positive Per Image (FPPI) uniformly sampled from $[10^{-2}, 10^0]$ \cite{dollar2012pedestrian}. More statistics about the annotated pedestrians can be viewed in Fig. \ref{fig:statistics}.
\setlength{\tabcolsep}{5.3pt}
\begin{table}[t]
\footnotesize
\centering
\begin{tabular}{l|ccccc}
\hline
Datasets& \#frame & \#ID & \#ped. & \#ped. w/ ID & \#ped. w/o ID \\
\hline
Train& 5,134 & 482 & 16,243 & 13,416 & 2,827 \\
Val.& 570 & 482 &1,805 & 1,491 & 314 \\
Test& 6,112 & 450 & 25,062 & 19,127 & 5,935 \\
\hline
\end{tabular}
\caption{Training/validation/testing split of the PRW dataset.}
\label{table:statistics_PRW}
\end{table}

{\bf Person Re-identification.}
A good re-ID system possesses two characteristics. First, all pedestrians are accurately localized within each frame, that is, 100\% recall and precision. Second, given a probe pedestrian, all instances of the same person captured by disjoint cameras are retrieved among the top-ranked results.

Re-ID is a 1:N search process. On the one hand, queries are produced by hand-drawn bounding boxes, as in practice, it takes acceptable time and effort for a user to draw a bounding box on the person-of-interest. For each ID, we randomly select one query under each camera. In total, we have 2,057 query images for the 450 IDs in the test set, averaging 4.57 (maximum 6) queries/ID. On the other hand, ``N'' denotes the database or gallery. A major difference between PRW and traditional re-ID datasets \cite{li2014deepreid, gray2007evaluating, zheng2009associating, liao2014open, das2014consistent, zheng2015scalable} is that the gallery in PRW varies with the settings of pedestrian detectors. Different detectors will produce galleries with different properties; even for the same type of detector, varying the detection threshold will yield galleries of different sizes. A good detector will be more likely to recall the person-of-interest while keeping the database small.

The IDs of the gallery boxes are determined by their intersection-over-union (IoU) scores with the ground truth boxes. In accordance to the practice in object detection, the detected boxes with IoU scores larger than 0.5 are assigned with an ID, while those with IoU less than 0.5 are determined as distractors \cite{zheng2015scalable}. Now, assume that we are given a query image $I$ and a gallery $\mathcal{G}$ generated by a specific detector. We calculate the similarity score between the query and all gallery boxes to obtain a ranking result. Following \cite{zheng2015scalable}, two metrics are used to evaluate person re-ID accuracy -- mean Average Precision (mAP), which is the mean across all queries' Average Precision (AP) and the rank-1, 10, 20 accuracy denoting the possibility to locate at least one true positive in the top-1, 10, 20 ranks.

Combining pedestrian detection, we plot mAP (or rank-1, rank-20 accuracy) against the average number of detected boxes per image to present the end-to-end re-ID performance. Conceptually, with few detection boxes per image, the detections are accurate but recall is low, so a small mAP is expected. When more boxes are detected, the gallery is filled with an increasing number of false positive detections, so mAP will first increase due to higher recall and then drop due to the influence of distractors.

\section{Base Components and Our Improvements}
\label{sec:components_and_improvements}

\subsection{Base Components in the End-to-End System}

{\bf Pedestrian detection.} Recent pedestrian detectors usually adopts the ``proposal+CNN'' approach \cite{tian2015deep, cai2015learning}. Instead of using objectness proposals such as selective search \cite{van2011segmentation}, hand-crafted pedestrian detectors are first applied to generate proposals. Since these weak detectors are discriminatively trained on pedestrians, it is possible to achieve good recall with very few proposals (in the order of 10). While RCNN is slow with $2000$ proposals, extracting CNN features from a small number of proposals is fast, so we use RCNN instead of the fast variant \cite{girshick2015fast}. Specifically, the feature for detection can be learnt through the RCNN framework by classifying each box into 2 classes, namely pedestrian and background. In this paper, three CNN architectures are tested: AlexNet \cite{krizhevsky2012imagenet}, VGGNet \cite{simonyan2014very} and ResNet \cite{he2015deep}.

{\bf Person re-identification.} We first describe some traditional methods. For image descriptors, we test 6 state-of-the-art methods, namly BoW \cite{zheng2015scalable}, LOMO \cite{liao2015person}, gBiCov \cite{ma2014covariance}, HistLBP \cite{xiong2014person}, SDALF \cite{farenzena2010person} and the IDE recognizer we propose in Section \ref{sec:improvements}. For metric learning, the 4 tested methods are KISSME \cite{kostinger2012large}, XQDA \cite{liao2015person}, DVR \cite{wang2014person} and DNS \cite{zhang2016learning}.

For CNN-based methods, it is pointed out in \cite{zheng2016survey} that the identification model outperforms the siamese model given sufficient training data per class. In this work, the training samples per ID consist of both hand-drawn and detected boxes, and the average number of training samples per ID is over 50. So we can readily adopt the identification CNN model. Note that the training data do not include background detections due to their imbalance large number compared with the boxes for each ID. We do not apply any data augmentation. During training, a CNN embedding is learned to discriminate different identities. During testing, features of the detected bounding box are extracted from FC7 (AlexNet) after RELU, following which Euclidean distance or learned metrics are used for similarity calculation. We name the descriptor as ID-discriminative Embedding (IDE). The implementation details of IDE can be viewed in \cite{zheng2016survey,zheng2017unlabeled}\footnote{\scriptsize{\url{github.com/zhunzhong07/IDE-baseline-Market-1501}}}.

\subsection{Proposed Improvements}
\label{sec:improvements}

\vspace{-0.2cm}
{\bf Cascaded fine-tuning strategy.} In \cite{zheng2016survey}, the IDE descriptor is fine-tuned using the Market-1501 dataset \cite{zheng2015scalable} on the ImageNet pre-trained model. In this paper, we name this descriptor as IDE$_{imgnet}$ and treat it as a competing method. For the proposed cascaded fine-tuning strategy, we insert another fine-tuning step in the generation process of  IDE$_{imgnet}$. That is, build on the ImageNet pre-trained model, we first train a 2-class recognition model using the detection data, \ie, to tell whether an image contains a pedestrian or not. Then, we train a 482-class recognition model using the training data of PRW. The two fine-tuning process which is called ``cascaded fine-tuning'', results in a new CNN embedding, denoted as  IDE$_{det}$. The two types of CNN embeddings are summarized below:

\vspace{-0.2cm}
\begin{itemize}
\item \textbf{IDE$_{imgnet}$}. The IDE model is directly fine-tuned on the ImageNet pre-trained CNN model. In what follows, when not specified, we use the term \textbf{IDE} to stand for \textbf{IDE$_{imgnet}$} for simplicity.
\vspace{-0.2cm}
\item \textbf{IDE$_{det}$}. With the ImageNet pre-trained CNN model, we first train an R-CNN model on PRW which is a two-class recognition task comprising of pedestrians and the background. Then, we fine-tune the R-CNN model with the IDE method, resulting in IDE$_{det}$.
\end{itemize}
\vspace{-0.2cm}

Through the cascaded fine-tuning strategy, the learned descriptor has ``seen'' more background training samples as well as more pedestrians (labeled as ``-2'') that are provided by the detection label of PRW. Therefore, the learned descriptor IDE$_{det}$ has improved discriminative ability to reduce the impact of false detections on the background. In the experiment, the performance of the two variants will be compared and insights will be drawn.

{\bf Confidence Weighted Similarity.} Previous works treat all gallery boxes as equal in estimating their similarity with the query. This results in a problem: when populated with false detections on the background (inevitable when gallery gets larger with the use of detectors), re-ID accuracy will drop with the gallery size \cite{zheng2015scalable}. This work proposes to address this problem by incorporating detection confidence into the similarity measurement. Intuitively, false positive detections will receive lower weights and will have reduced impact on re-ID accuracy. Specifically, detector confidences of all gallery boxes are linearly normalized to $[0, 1]$ in a global manner. Then, the cosine distance between two descriptors are calculated, before multiplying the normalized confidence. Note that the \textbf{IDE feature is extracted from FC7 after RELU in AlexNet, so there are no negative entries in the IDE vector.} The cosine distance remains non-negative with IDE vectors, and is compatible with the detection scores. Currently, this baseline method supports cosine (Euclidean) distance between descriptors, but in future works, more sophisticated weightings corresponding to metric learning methods may also be considered, which should be a novel research direction in person re-ID.

\makeatother
\begin{figure} [t]
\centering
\subfigure[Recall, IoU$>$0.5]{\label{fig:state-of-the-art-iLIDS}%
\includegraphics[width=1.6in]{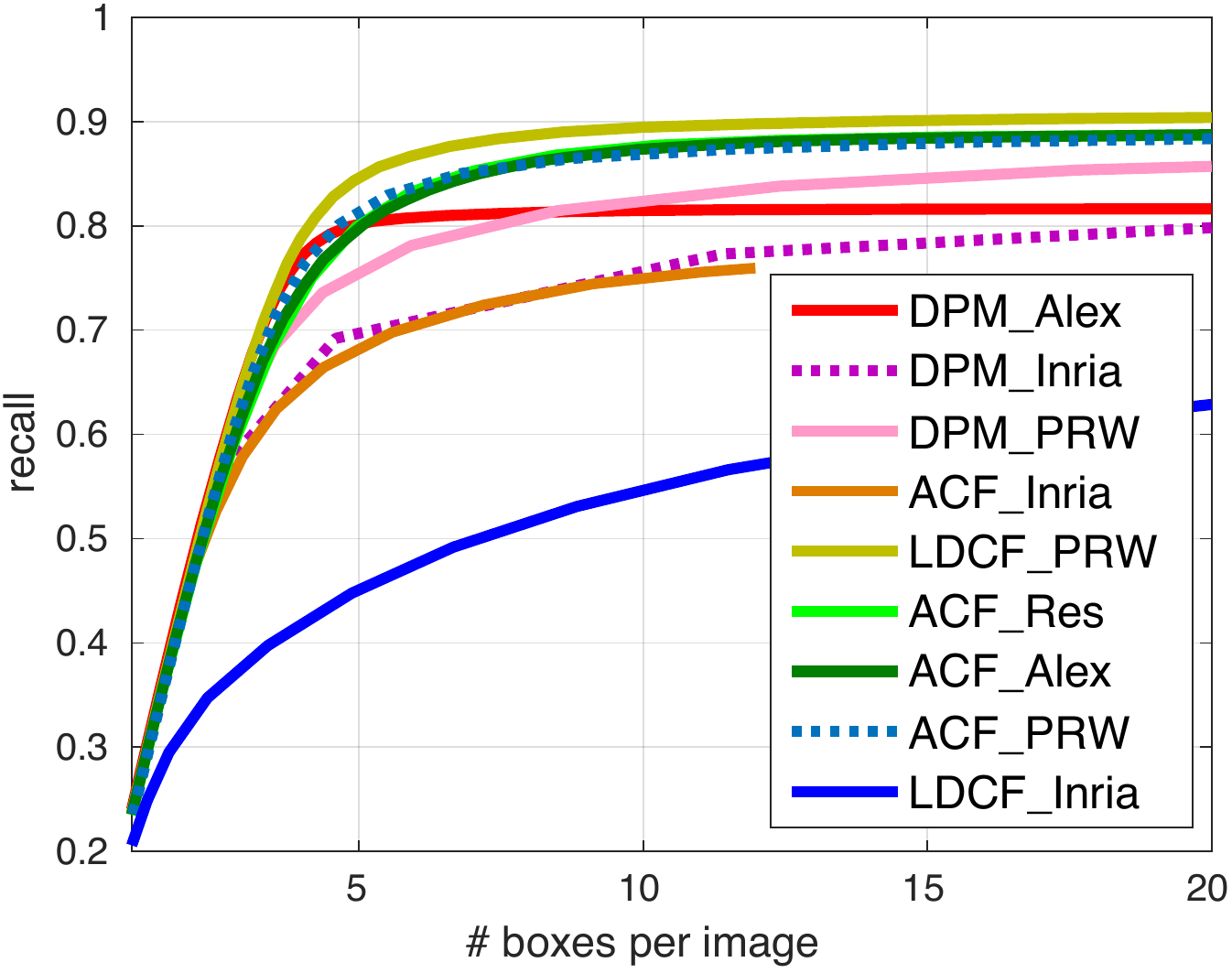}}
 \subfigure[Recall, IoU$>$0.7]{\label{fig:state-of-the-art-PRID}%
\includegraphics[width=1.6in]{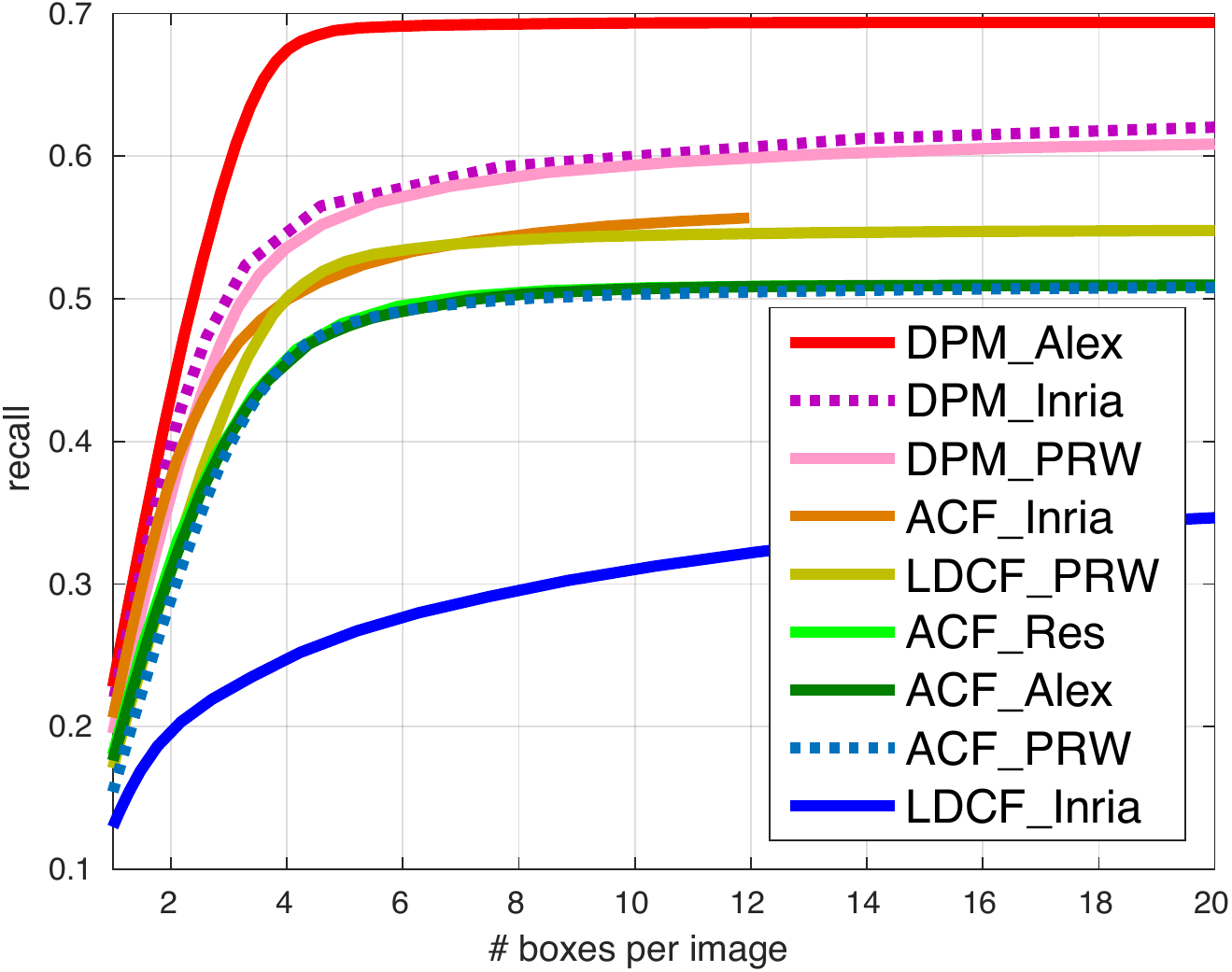}}
\caption{Detection recall at two IoU criteria. ``Inria'' and ``PRW'' denote models trained on INRIA \cite{dalal2005histograms} and the proposed PRW datasets, respectively. ``Alex'' and ``Res'' denote RCNN models fine-tuned with AlexNet \cite{krizhevsky2012imagenet} and ResidualNet \cite{he2015deep}, respectively. For IoU$>$0.7, we use warm colors for detectors with higher AP, and cold colors for bad detectors. Best viewed in color.}
\label{fig:recall}
\end {figure}

\makeatother
\begin{figure} [t]
\centering
\subfigure[Precision-Recall, IoU$>$0.5]{\label{fig:pr05}%
\includegraphics[width=1.6in]{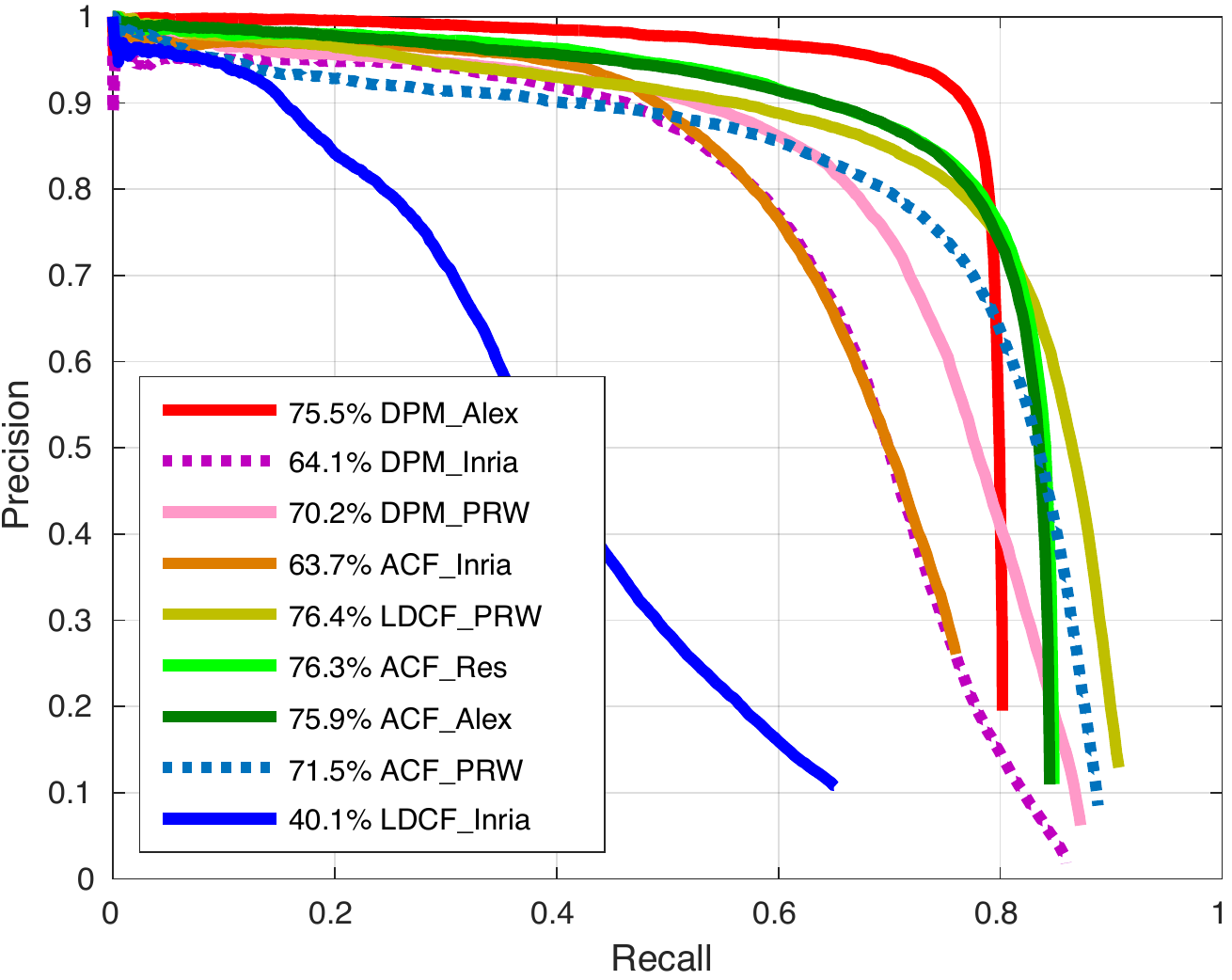}}
 \subfigure[Precision-Recall, IoU$>$0.7]{\label{fig:pr07}%
\includegraphics[width=1.6in]{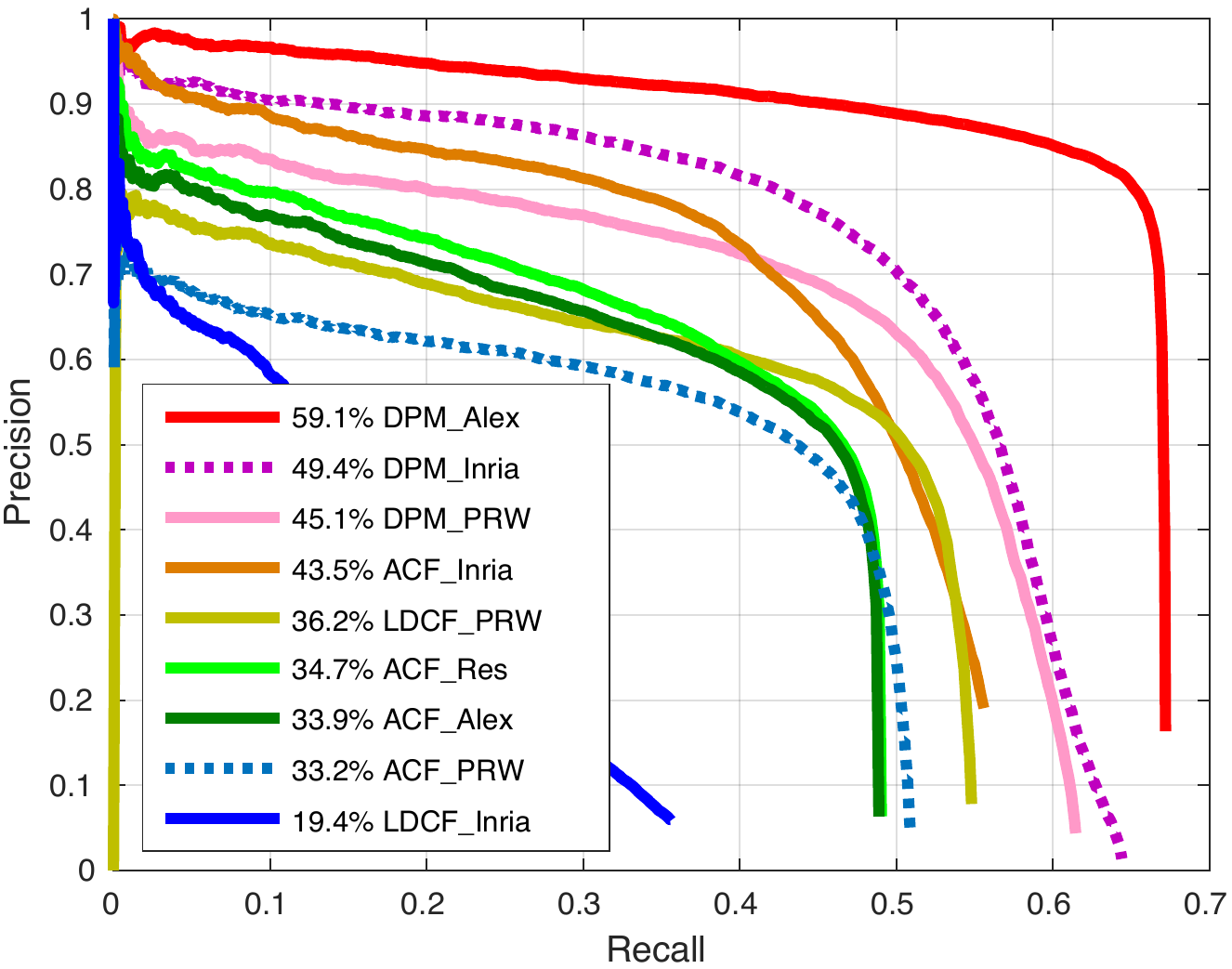}}
\caption{Precision-recall curves at two IoU criteria. Detector legends are the same as Fig.~\ref{fig:recall} (Best viewed in color). The Average Precision number is shown before the name of each method. }
\label{fig:pr}
\end {figure}

\section{Experiments}
\label{sec:experiment}

\subsection{Evaluation of Pedestrian Detection}
First, we evaluate the detection recall of several important detection models on PRW. This serves as an important reference to the effectiveness of proposals for RCNN based methods. These models include Deformable Part Model (DPM) \cite{felzenszwalb2010object}, Aggregated Channel Features (ACF) \cite{dollar2014fast} and Locally Decorrelated Channel Features (LDCF) \cite{nam2014local}. We also test their respective RCNN versions. We retrain these models on the PRW training set and plot detection recall against average number of detection boxes per image on the testing set. The results are shown in Fig.~\ref{fig:recall}. It is observed that recall is relatively low for the off-the-shelf detectors. After being retrained on PED1K dataset, LDCF yields recall of 89.3\% on 11.2 proposals per image; ACF produces recall of 88.81\% with 34.5 proposals per image; DPM will have a recall of 86.81\% with 32.3 proposals on average. These results are collected under IoU $>0.5$. When IoU increases to 0.7, detector recalls deteriorate significantly. In all, recall for the best detectors reaches around $90\%$ for IoU $> 0.5$, and around $60\%$ for IoU $> 0.7$.

The detection methods without RCNN mentioned above are used as proposals, and are subsequently coupled with RCNN based models. Specifically, we fine tune RCNN with three CNN models -- AlexNet (Alex) \cite{krizhevsky2012imagenet}, VGGNet (VGG)\cite{simonyan2014very}, and ResidualNet (Res) \cite{he2015deep}. Additionally, we report results of the Histogram of Oriented Gradient (HOG) \cite{dalal2005histograms}. True positives are defined by IoU $> 0.5$ or IoU $> 0.7$. We report both the Average Precision (AP) and Log Average Miss Rate (MR). Experimental results are presented in Fig.~\ref{fig:pr}. As IoU increases, detector performance deteriorates significantly which is observed in \cite{dollar2012pedestrian}. Under IoU $> 0.7$, the best detector is DPM+AlexNet, having an AP of 59.1\%, which is +9.7\% higher than the second best detector. The reason that DPM has robust performance under larger IoU is that it consists of multiple components (parts) which adapts well to pedestrian deformation, while channel feature based methods typically set aspect ratio types that are less robust to target variations. In both detection recall and detection accuracy experiments, we find that detector rankings are different from IoU $> 0.5$ to IoU $> 0.7$. With respect to detection time, it takes 2.7s, 1.4s, and 6.5s on average on a 1080$\times$1920 frame for ACF, LDCF and DPM, respectively, using MATLAB 2015B on a machine with 16GB memory, K40 GPU and Intel i7-4770 Processor. RCNN requires 0.2s for 20 proposals.

%

From these benchmarking results, it is shown that the usage of RCNN effectively increases detection performance given a proposal type. For example, when using ACF as proposal, the inclusion of AlexNet increases AP from 74.16\% to 76.23\% (+2.07\%). Further, when different CNN models are used for a given proposal, we find that ResidualNet outperforms the others in general: AP of ResNet is +0.41\% higher than AlexNet.

\setlength{\tabcolsep}{2.0pt}
\begin{table}[t]
\scriptsize
\centering
\begin{tabular}{l|l|ccc|ccc|ccc}
    \cline{2-10}
\hline
    \multirow{2}{*}{Detector} &
    \multirow{2}{*}{Recognizer} &
    \multicolumn{3}{c|}{{\#detection=3}}&
    \multicolumn{3}{c|}{{\#detection=5}} &
    \multicolumn{3}{c}{{\#detection=10}} \\
& &mAP & r1 & r20 & mAP & r1 & r20 & mAP & r1 & r20\\

\hline
 DPM&BOW& 8.9 & 30.4 & 58.3 & 9.7 & 31.1& 58.6&9.6&30.5&57.7 \\
 DPM&IDE& 12.7 & 37.2 & 72.2 & 13.7 & 36.9 & 72.1& 13.7&36.6&70.8\\
 DPM&IDE$_{det}$& 17.2&45.9  & 77.9 & 18.8 & 45.9 &77.4&19.2&45.7&76.0\\
 DPM-Alex&{SDALF+Kiss.}& 12.0 & 32.6 &63.8 & 13.0 &32.5 &63.4&12.4&31.8& \\
 DPM-Alex&{LOMO+XQ.}& 13.4 & 34.9 &66.5 & 13.0 &34.1 &64.0&12.4&33.6&62.5 \\
 DPM-Alex&{HistLBP+DNS}& 14.1 & 36.8 &70.0 & 13.6 &35.9 &67.8&12.7&35.0&65.7 \\
 DPM-Alex&IDE& 15.1 & 38.8 & 74.1 & 14.8 &37.6 &71.4&14.1& 36.9&69.8\\
 DPM-Alex&IDE$_{det}$& \bf20.2 & \bf48.2 & 78.1 &20.3  &47.4&77.1&19.9&47.2& 76.4 \\
\rowcolor{mygray}
 DPM-Alex&IDE$_{det}$+CWS& 20.0 & \bf48.2 & \bf78.8 & \bf20.5 &\bf48.3 &\bf78.8&\bf20.5&\bf48.3& \bf78.8\\
\hline
 ACF&LOMO+XQ.& 10.5& 31.5 & 61.6& 10.5&30.9 &59.5 &9.7&29.7&57.4\\
 ACF&gBiCov+Kiss.& 9.8 & 31.1 & 60.1 & 9.9 & 30.3&58.3&9.0&29.0& 55.9\\
 ACF&IDE$_{det}$&16.6  & 44.8 & 75.9 & 17.5 &43.8 &76.0&17.0&42.9&74.5 \\
 ACF-Res&IDE& 12.4  & 35.0 & 70.4 & 12.5 & 33.8&68.6&11.5&33.0& 66.7\\
 ACF-Alex&LOMO+XQ.& 10.5& 31.8 & 60.7& 10.3&30.6 & 59.4&9.5&29.6&57.1\\
 ACF-Alex&IDE$_{det}$& 17.0 & 45.2 & 76.6 &17.5  & 43.6&75.1&16.6&42.7& 73.7\\
\rowcolor{mygray}
 ACF-Alex&IDE$_{det}$+CWS& 17.0 & 45.2 & 76.8& 17.8 & 45.2 &76.8&17.8&45.2&76.8\\
\hline
 LDCF&BoW& 8.2 & 30.1 & 56.9 &9.1  &29.8 &57.0 &8.3&28.3&55.3\\
 LDCF&LOMO+XQ.& 11.2 & 31.6 & 62.9 & 11.0 & 31.1 &62.2 &10.1&29.6&58.6\\
 LDCF&gBiCov+Kiss.& 9.5 & 30.7 & 58.8 & 9.6 &30.1 &58.4&8.8&28.7& 56.7\\
 LDCF&IDE& 12.7 & 35.3 & 70.1 & 34.4 & 13.1&69.4&12.2&33.1&68.0 \\
 LDCF&IDE$_{det}$& 17.5 & 45.3 & 76.2 & 18.3 &44.6 & 75.6&17.7&43.8&74.3\\
\rowcolor{mygray}
 LDCF&IDE$_{det}$+CWS& 17.5 & 45.5& 76.3 & 18.3 & 45.5 &76.4&18.3& 45.5&76.4\\
\hline
\end{tabular}
\caption{Benchmarking results of various combinations of detectors and recognizers on the PRW dataset.}
\label{table:reid_summary}
\end{table}
Similar to the performance of proposals, under IoU $>$ 0.7, detector performance deteriorates significantly which is observed in \cite{dollar2012pedestrian}.
For example, LDCF yields the highest recall under IoU $>$ 0.5, while it only ranks 4th under IoU $>$ 0.7. When measured under IoU $>$ 0.7, the DPM detectors are superior, probably because DPM deals with object deformation by detecting parts and adapts well to PRW where pedestrians have diverse aspect ratios (see Fig. \ref{fig:aspect_ratio}).

\subsection{Evaluation of Person Re-identification}
We benchmark the performance of some recent descriptors and distance metrics on the PRW dataset. Various types of detectors are used -- DPM, ACF, LDCF and their related RCNN methods. The descriptors we have tested include the Bag-of-Words vector \cite{zheng2015scalable}, the IDE descriptor described in Section \ref{sec:improvements}, SDALF \cite{farenzena2010person}, LOMO \cite{liao2015person}, HistLBP \cite{xiong2014person},  and gBiCov \cite{ma2014covariance}. The used metric learning methods include Kissme \cite{kostinger2012large}, XQDA \cite{liao2015person}, and the newly proposed DNS \cite{zhang2016learning}. The results are shown in Fig.~\ref{fig:detector_on_reid} and Table \ref{table:reid_summary}.

The unsupervised descriptor BoW yields decent performance on PRW dataset: around 10\% in mAP and 30\% in rank-1 accuracy. Improvements can be found when metric learning methods are employed. for example, when coupling SDALF and Kissme, mAP increases to 12.0\% and 32.6\% in mAP and rank-1 accuracy, respectively. We observe that for hand-crafted features, ``HistLBP+DNS'' outperforms others when built on the DPM-AlexNet detector. These results generally agree with observations in prior works \cite{zhang2016learning}. We conjecture that given a fixed detector, re-ID accuracy will display similar trends as prior studies \cite{xiong2014person,zhang2016learning,li2014deepreid}. The IDE descriptor yields significantly higher accuracy compared with the others. For example, IDE$_{det}$ exceeds ``HistLBP+DNS'' by +6.2\% in mAP when on average 3 bounding boxes are detected per image. This validates the effectiveness of the CNN-based descriptor. When different detectors are employed, detectors with higher AP under IoU $>$0.7 are generally beneficial towards higher overall re-ID accuracy.

The number of detected bounding boxes per image also has an impact on re-ID performance. When too few (\emph{e.g.,} 2) bounding boxes are detected, it is highly possible that our person-of-interest is not detected, so the overall re-ID accuracy can be compromised. But when too many bounding boxes are detected, distractors may exert negative influence on the re-ID accuracy, so accuracy will slowly drop as the number of bounding boxes per image increases (Fig. \ref{fig:detector_on_reid}). Nevertheless, one thing we should keep in mind is that with more bounding boxes, the timings for person retrieval also increase. Currently most works do not consider retrieval efficiency due to the small volume of the gallery.  PRW, on the other hand, may produce over 100k bounding boxes, so efficiency may become an important issue in future research.

\subsection{Impact of detectors on re-identification}

\makeatother
\begin{figure} [t]
\centering
\subfigure[BOW, mAP]{\label{fig:mAP_BOW}%
\includegraphics[width=1.0in]{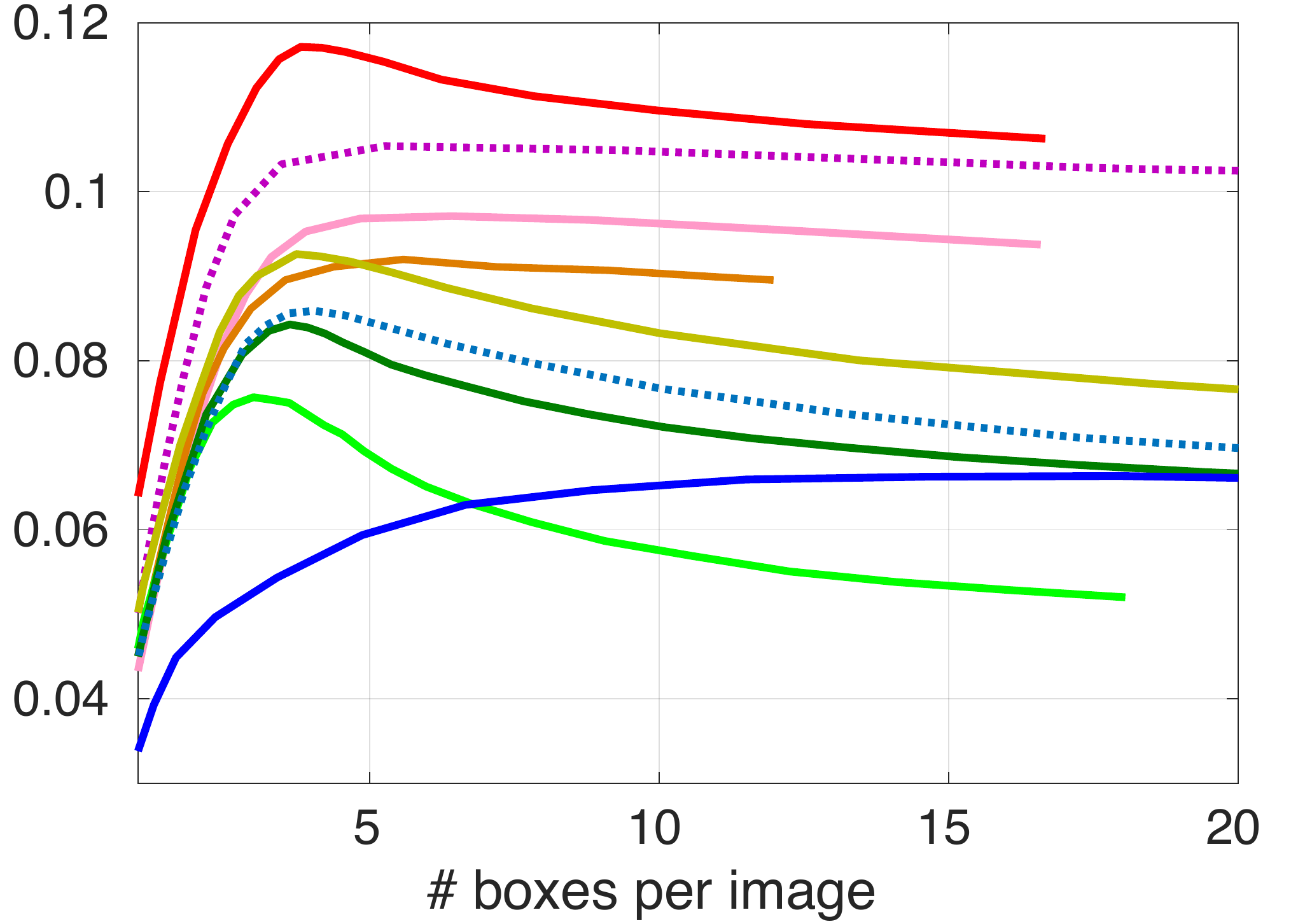}}
\hspace{0.in}
 \subfigure[BOW, Rank-1]{\label{fig:rank1_BOW}%
\includegraphics[width=1.05in]{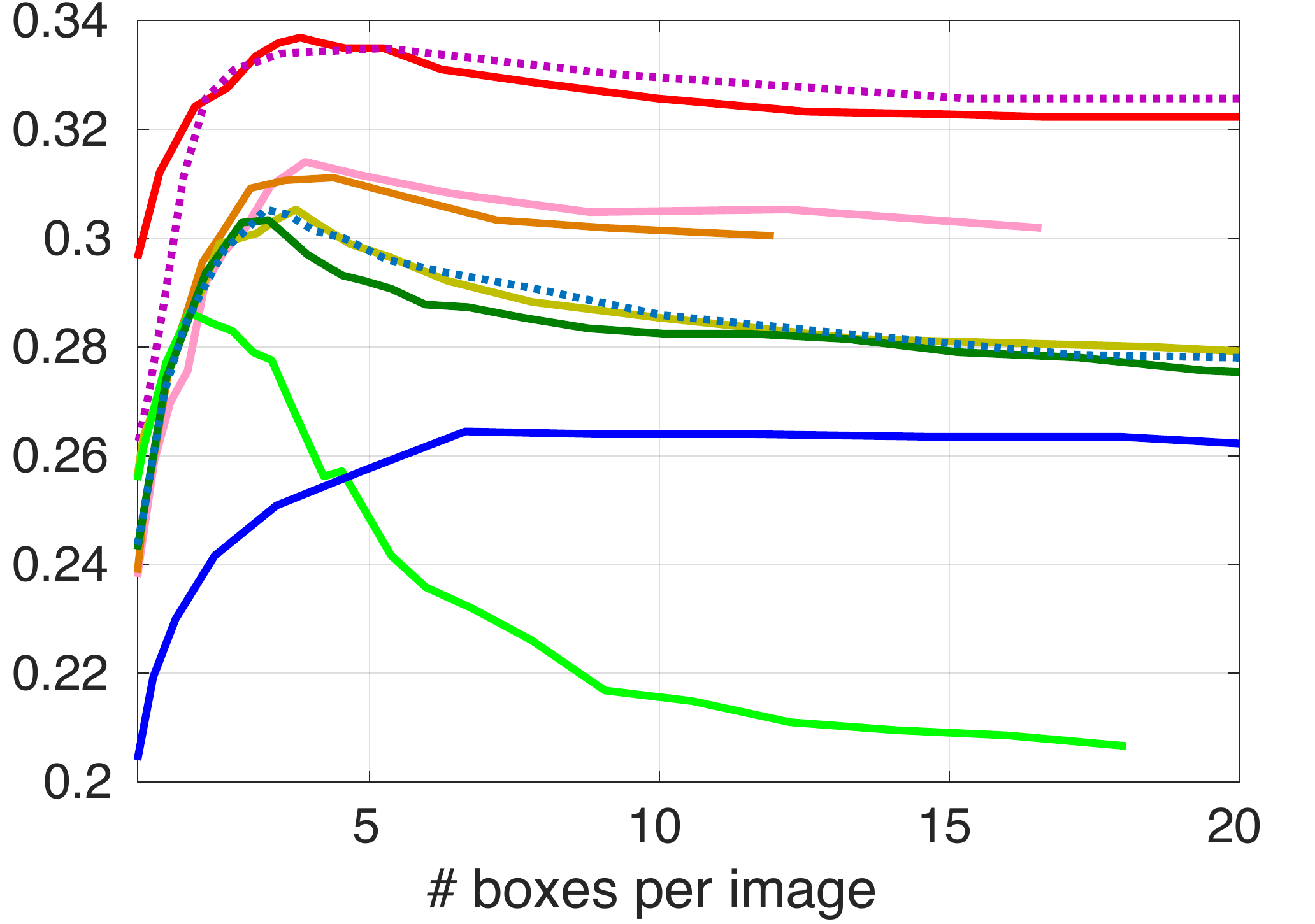}}
\hspace{0.in}
 \subfigure[BOW, Rank-20]{\label{fig:rank20_BOW}%
\includegraphics[width=1.05in]{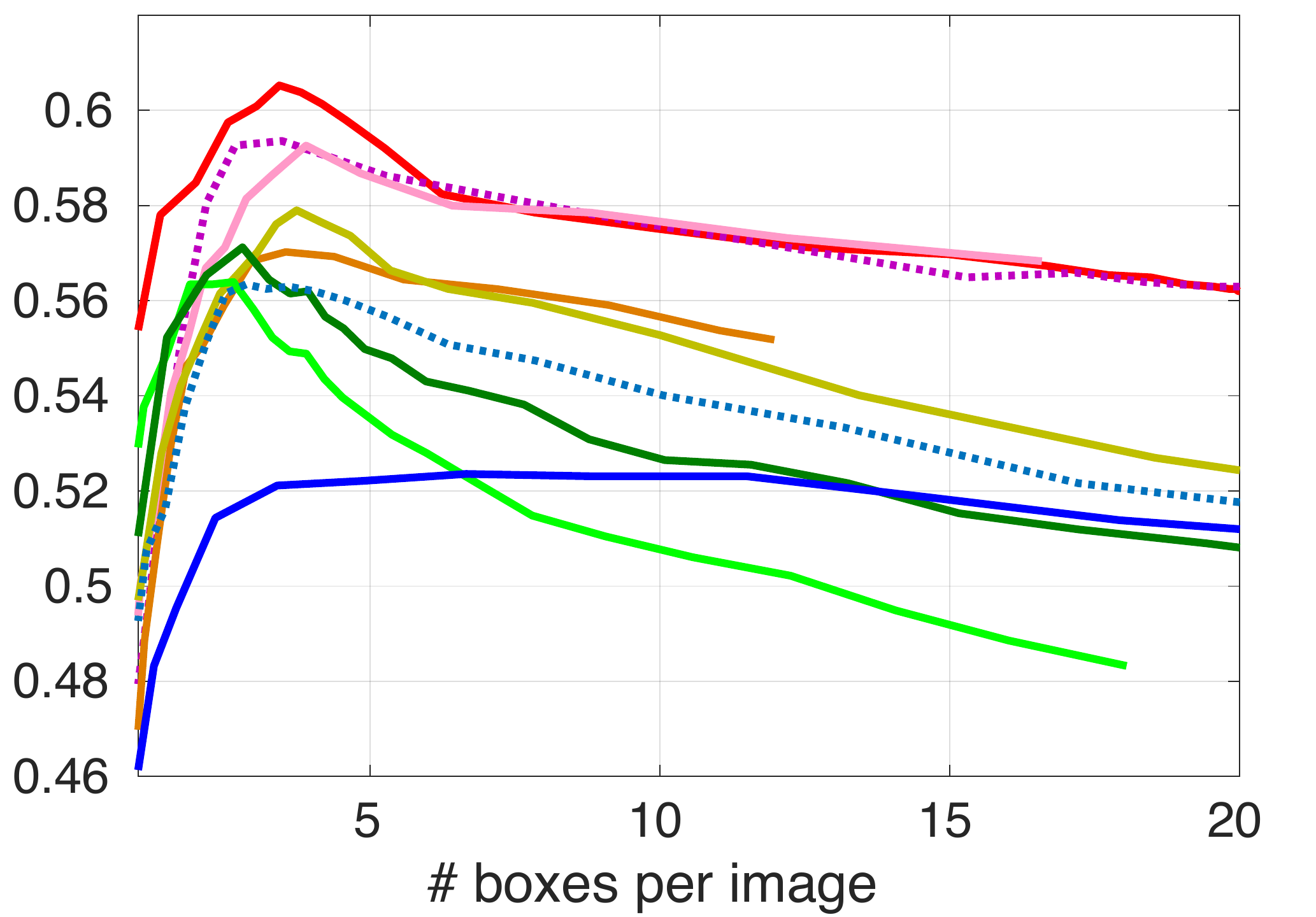}}
 \subfigure[IDE, mAP]{\label{fig:state-of-the-art-PRID1}%
\includegraphics[width=1.05in]{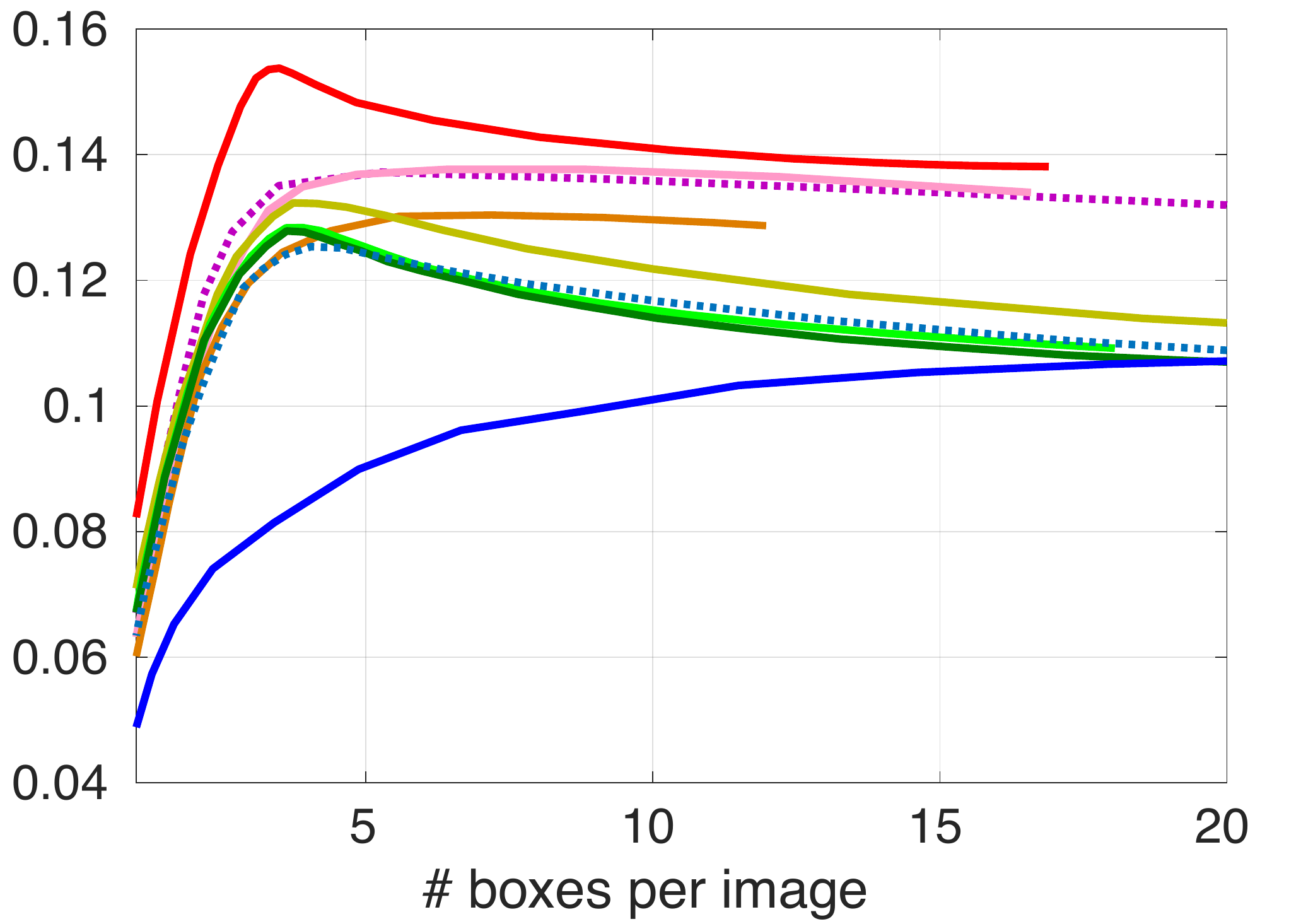}}
\hspace{0.in}
 \subfigure[IDE, Rank-1]{\label{fig:state-of-the-art-PRID2}%
\includegraphics[width=1.05in]{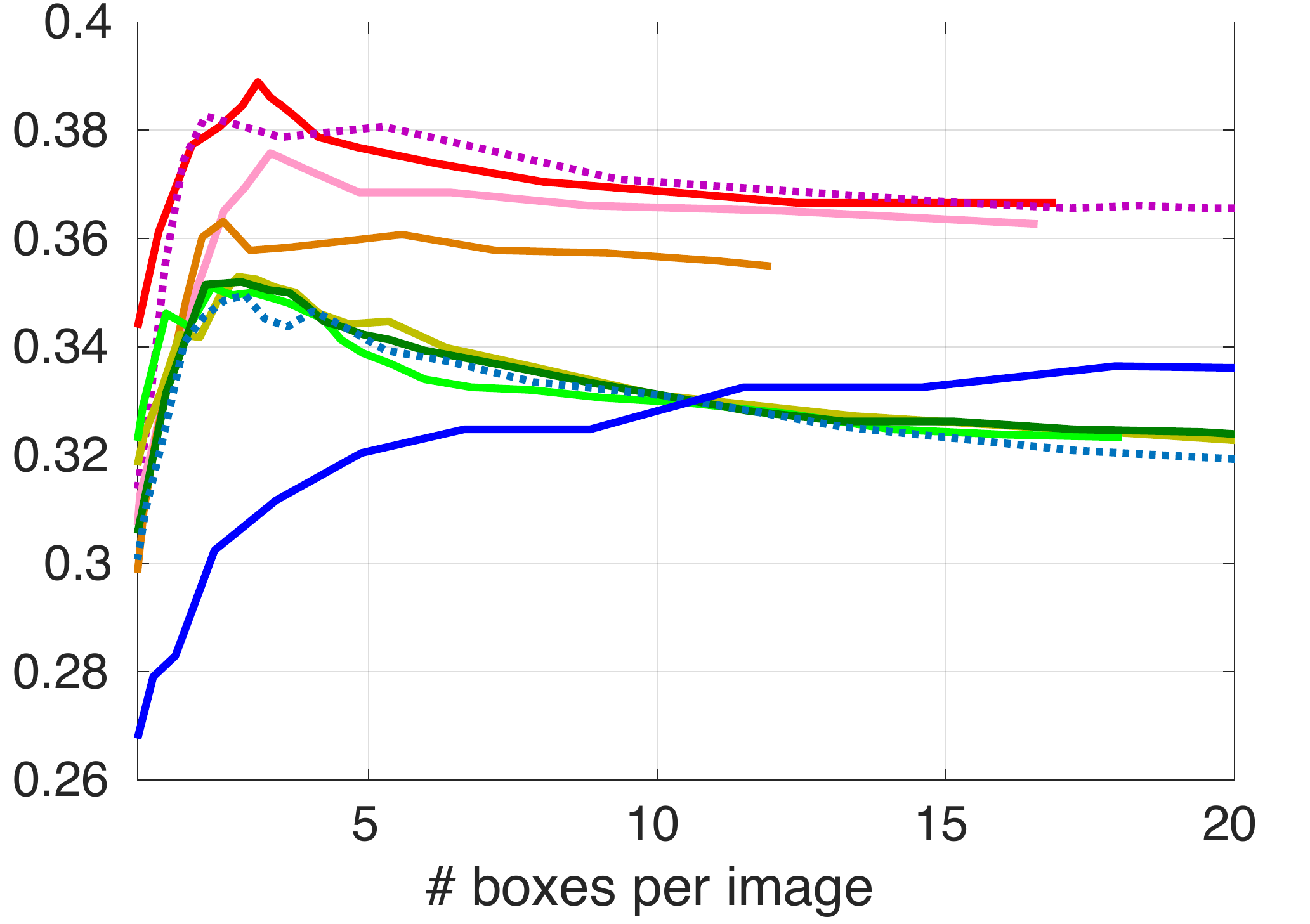}}
\hspace{0.in}
 \subfigure[IDE, Rank-20]{\label{fig:state-of-the-art-PRID3}%
\includegraphics[width=1.05in]{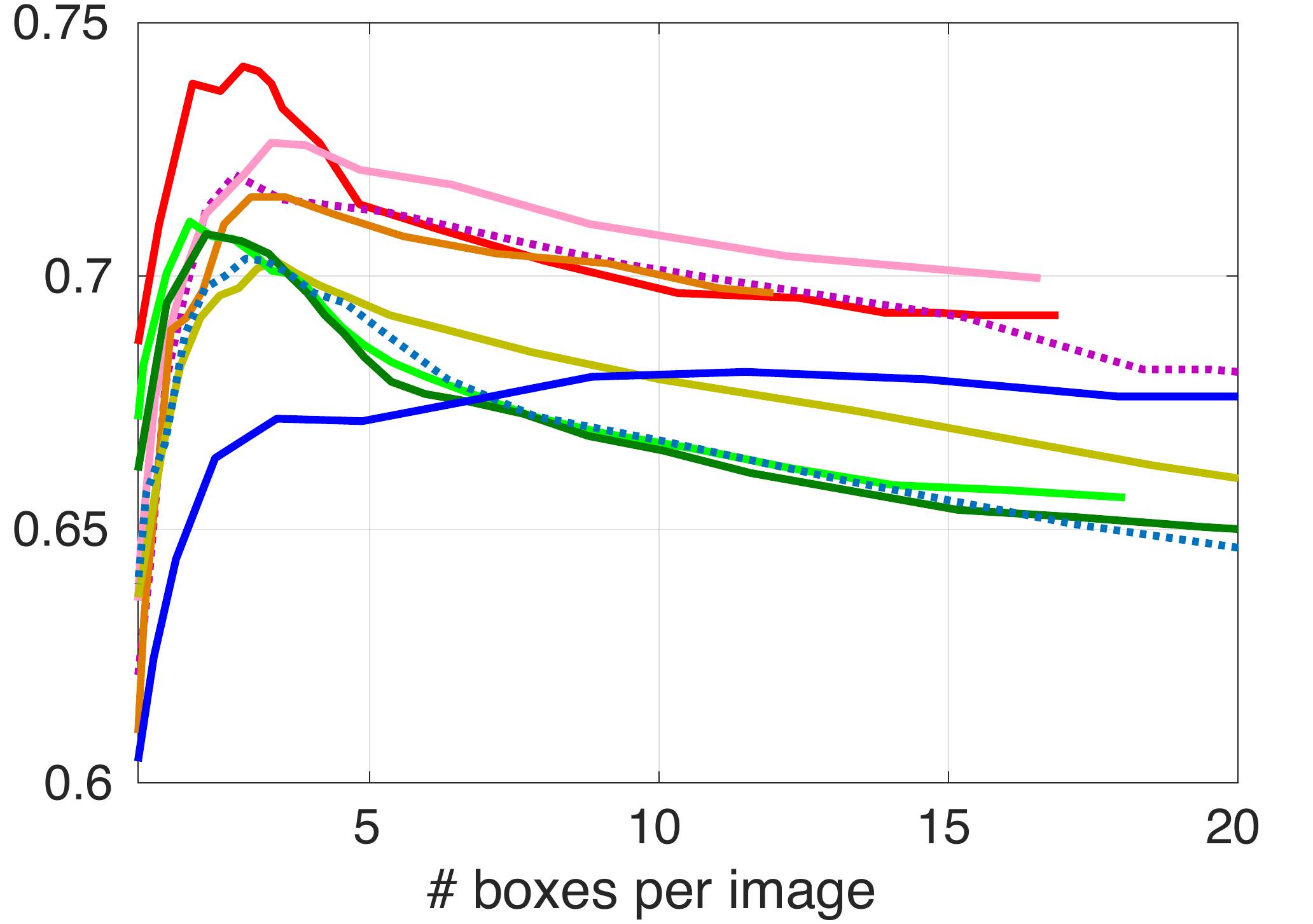}}
 \subfigure[LOMO, mAP]{\label{fig:state-of-the-art-PRID4}%
\includegraphics[width=1.05in]{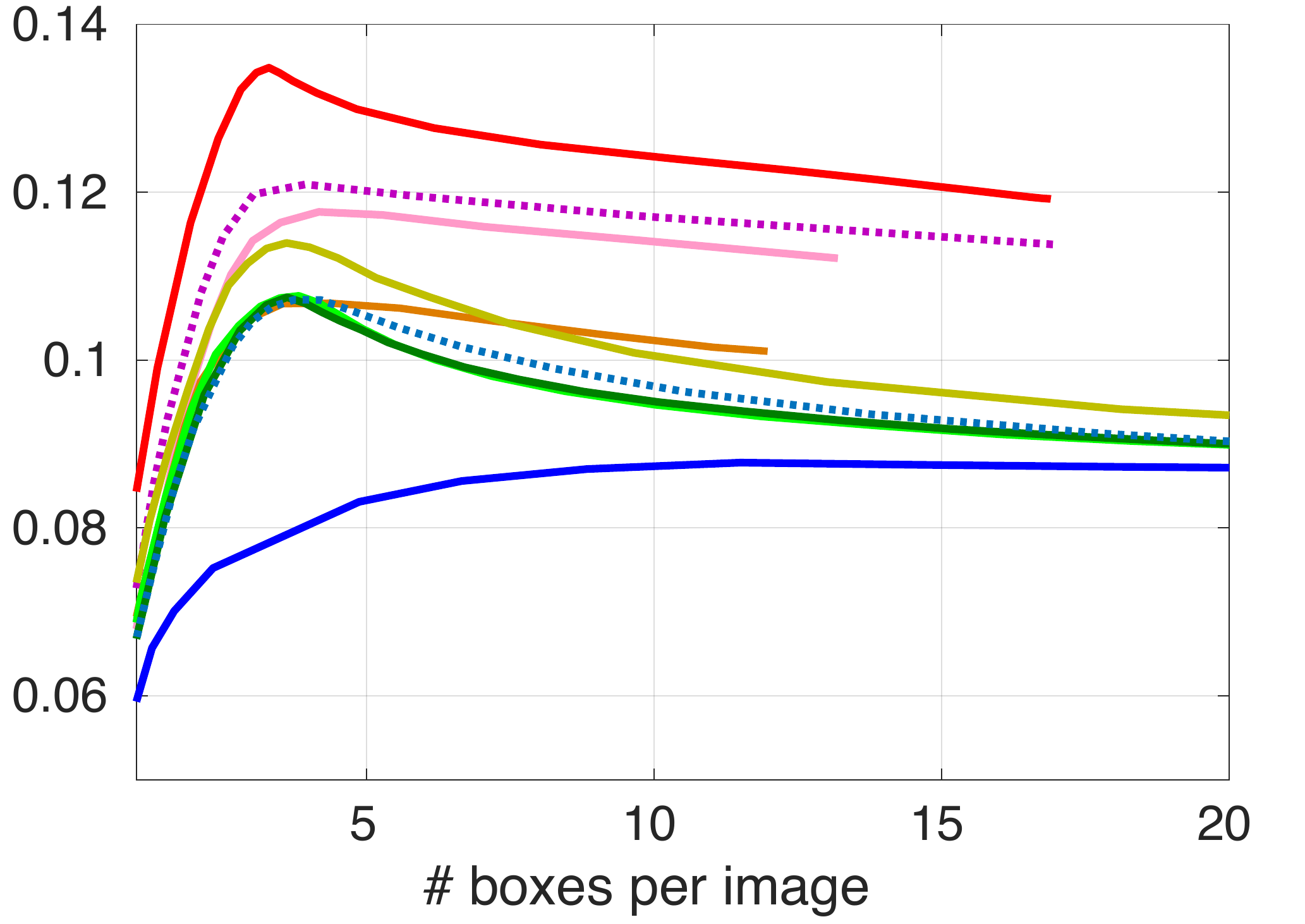}}
\hspace{0.in}
 \subfigure[LOMO, Rank-1]{\label{fig:state-of-the-art-PRID5}%
\includegraphics[width=1.05in]{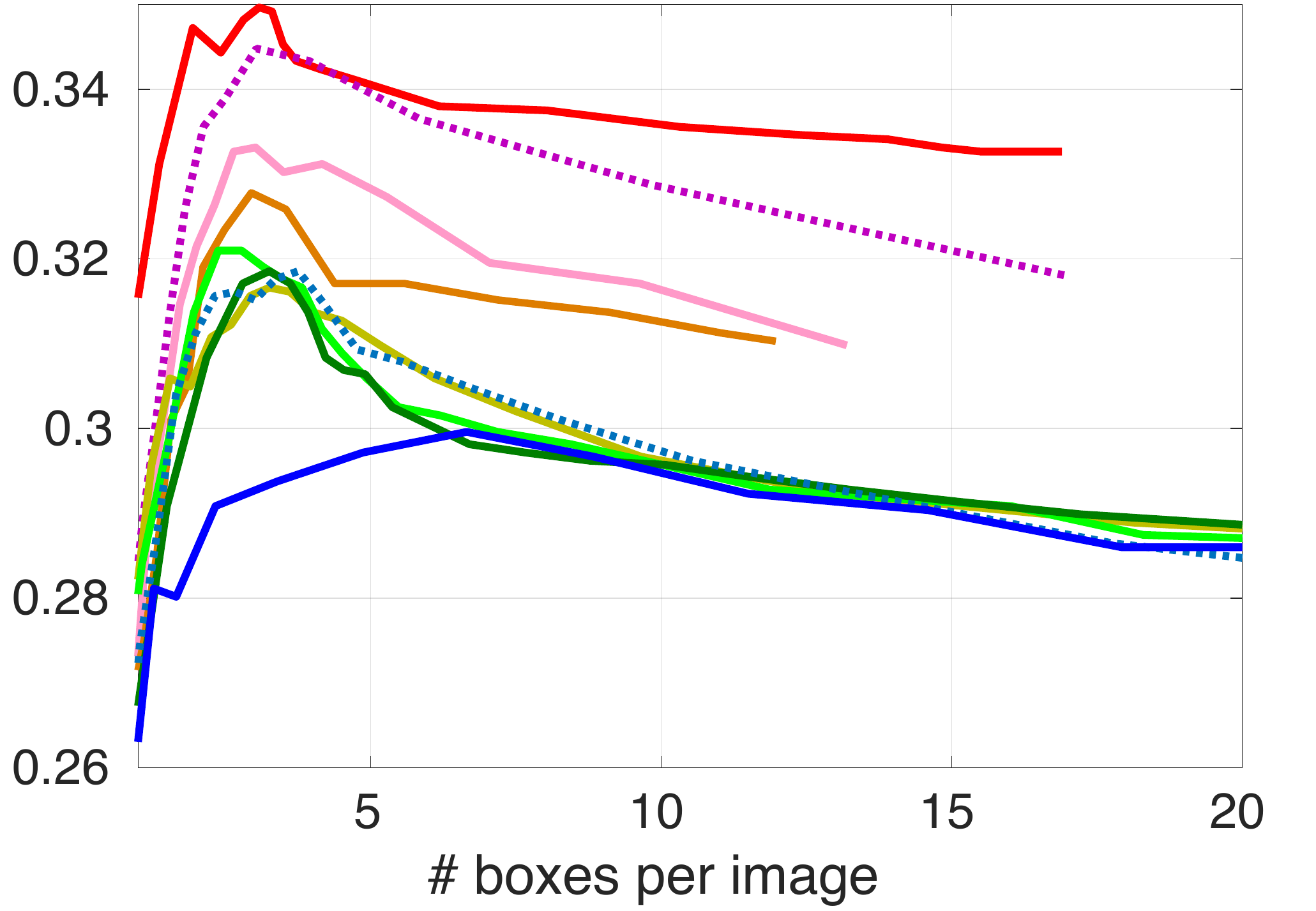}}
\hspace{0.in}
 \subfigure[LOMO, Rank-20]{\label{fig:state-of-the-art-PRID6}%
\includegraphics[width=1.05in]{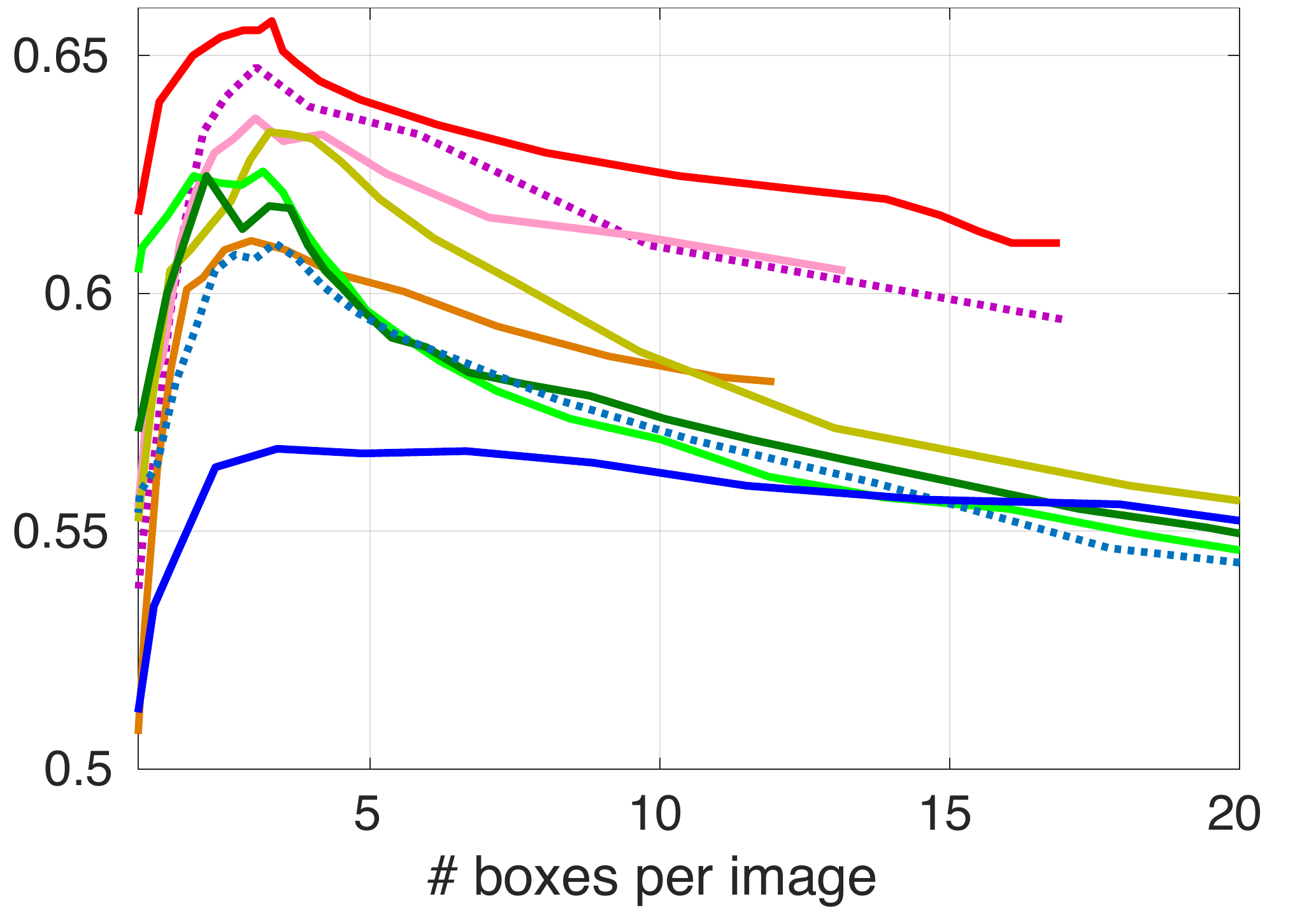}}
\caption{Re-id accuracy (mAP, rank-1 accuracy, and rank-20 accuracy) with 9 detectors and 3 recognizers. Detector legends are the same as Fig.~\ref{fig:recall} and Fig.~\ref{fig:pr}. Given a recognizer, we find that the performance of overall re-ID accuracy is more consistent with detection accuracy under IoU$>$0.7 than IoU$>$0.5, which suggests that IoU$>$0.7 is a better criterion for detector evaluation under the application of re-ID.}
\label{fig:detector_on_reid}
\end {figure}

{\bf Criteria for detector assessment.}
How does the detector performance affect re-ID? This is a critical question in an end-to-end re-id system. Broadly speaking, a better detector would result in a higher re-id accuracy. So how to assess detector quality in the scenario of person re-ID? When only considering pedestrian detection, the community uses AP or MR defined under IOU $>$ 0.5. In this paper, we argue that, apart from providing high recall and precision, it is of crucial importance that a detector give good localization results. Specifically, we find that IoU  $>$ 0.7 is a more effective criteria than IoU  $>$ 0.5 for detection evaluation in the scenario of person re-ID, which is the third contribution of this work.

 To find how re-ID accuracy varies with detector performance, we systematically test 9 detectors (as described in Fig.~\ref{fig:recall} and Fig.~\ref{fig:pr}) and 3 recognizers. The 3 recognizers are: 1) $5,600$-dimensional Bag-of-Words (BoW) descriptor \cite{zheng2015scalable}, the state-of-the-art unsupervised descriptor, 2) $4,096$-dimensional CNN embedding feature described in Section \ref{sec:improvements} using AlexNet, and 3) LOMO+XQDA \cite{liao2015person}, a state-of-the-art supervised recognizer. From the results in Fig. \ref{fig:detector_on_reid} and Table \ref{table:reid_summary}, a key finding is that {\em given a recognizer, the re-ID performance is consistent with detector performance evaluated using the IoU $> 0.7$ criterion}. In fact, if we use the IoU $> 0.5$ criterion as most commonly employed in pedestrian detection, our study shows that the detector rankings do not have accurate predictions on re-ID accuracy. Specifically, while the ``DPM\_Alex'' detector ranks 4th in average precision ($75.5\%$) with the IoU $> 0.5$ rule, it enables superior re-ID performance which is suggested in its top ranking under IoU $> 0.7$. The same observations hold for the other 8 detectors. This conclusion can be attributed to the explanation that under normal circumstances, a better localization result will enable more accurate matching between the query and gallery boxes. As an insight from this observation, when a pool of detectors is available in a practical person re-ID system, a good way for choosing the optimal one is to rank the detectors according to their performance under IoU $> 0.7$. In addition, recent research on partial person re-ID \cite{zheng2015partial} may be a possible solution to the problem of misalignment.

\makeatother
\begin{figure} [t]
\centering
\subfigure[mAP, 3 boxes/img]{\label{fig:mAP_3}%
\includegraphics[width=1.6in]{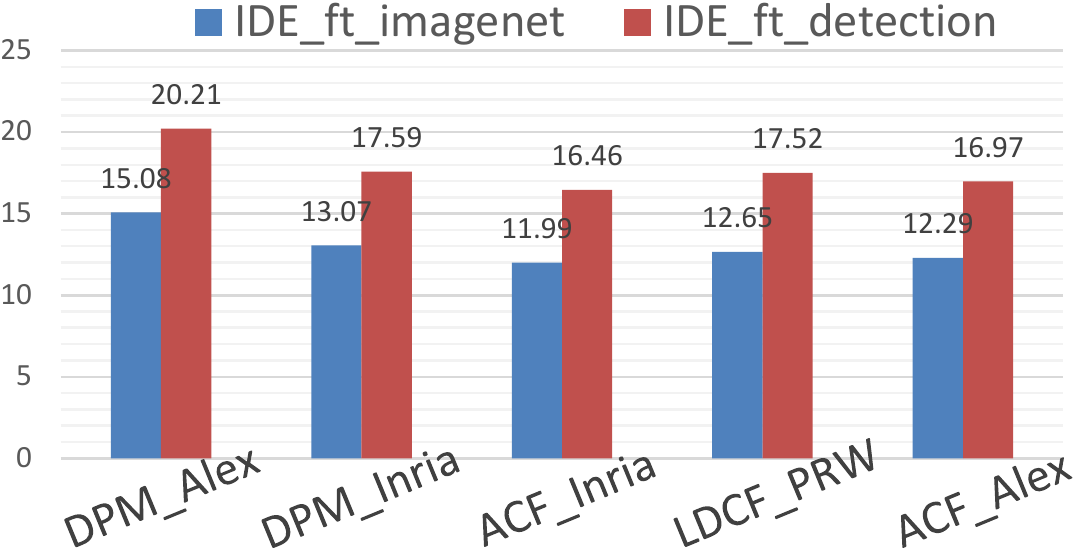}}
\hspace{0.in}
 \subfigure[mAP, 5 boxes/img]{\label{fig:mAP_5}%
\includegraphics[width=1.6in]{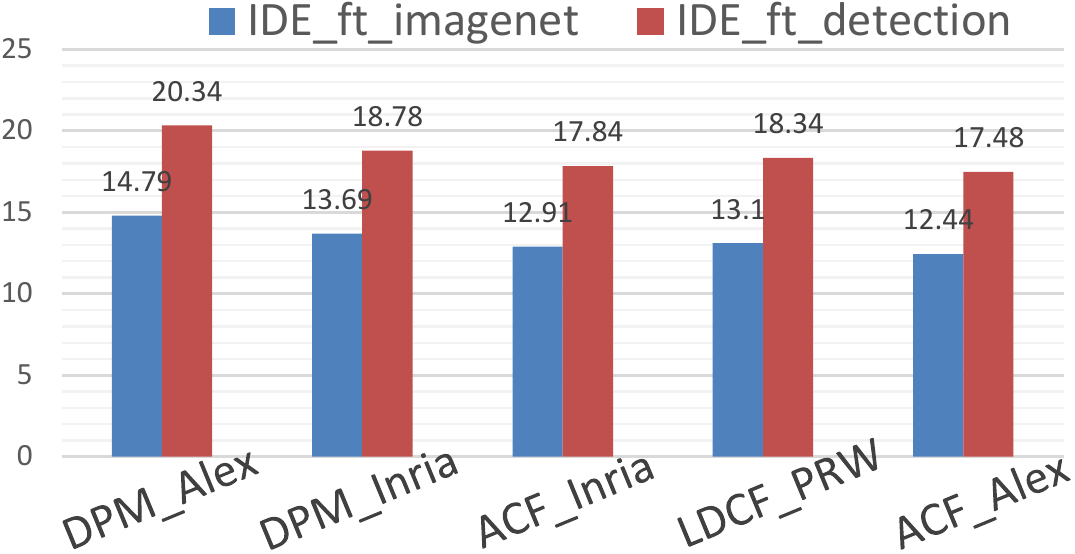}}
\subfigure[rank-1 acc., 3 boxes/img]{\label{fig:rank1_3}%
\includegraphics[width=1.6in]{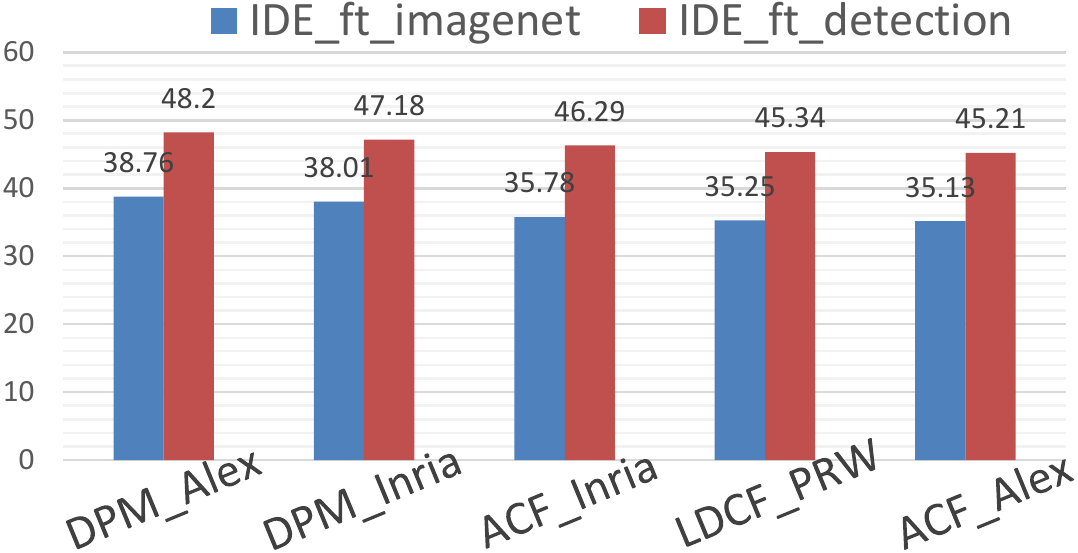}}
\hspace{0.in}
 \subfigure[rank-1 acc., 5 boxes/img]{\label{fig:rank1_5}%
\includegraphics[width=1.6in]{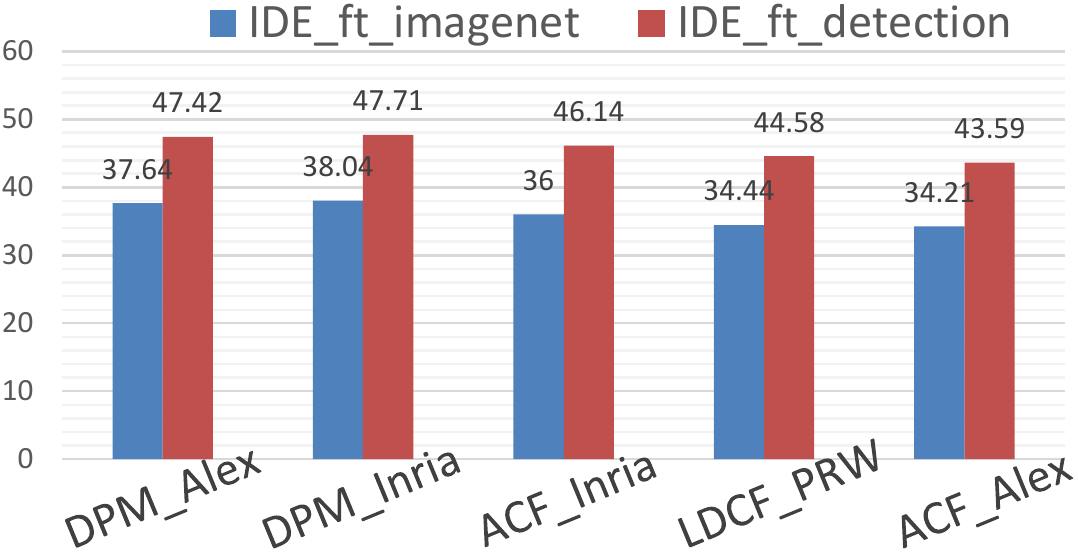}}
\caption{Plots of mAP and rank-1 accuracy using two variants of the IDE with 5 detectors. Fine-tuning on the pedestrian-background detection model improves over fine-tuning on the Imagenet model, proving the effectiveness of the proposed cascaded fine-tuning method.}
\label{fig:IDE}
\end {figure}

{\bf Effectiveness of cascade fine-tuning.} This paper introduces two IDE variants. For the first variant, we fine-tune IDE directly from AlexNet pre-trained on ImageNet, denoted as IDE$_{imgnet}$. For the second variant, we first fine-tune a pedestrian detection model (2 classes, pedestrian and background) from AlexNet pre-trained on ImageNet, and then we further fine tune it using the identification model on PRW. We denote the second variant as IDE$_{det}$, which is the learned embedding by the cascaded fine-tuning method. Experimental results related to the IDE variants are presented in Table \ref{table:reid_summary} and Fig.~\ref{fig:IDE}.

\makeatother
\begin{figure} [t]
\centering
 \subfigure[rank-1, CWS]{\label{fig:rank1_CWS}%
\includegraphics[width=1.05in]{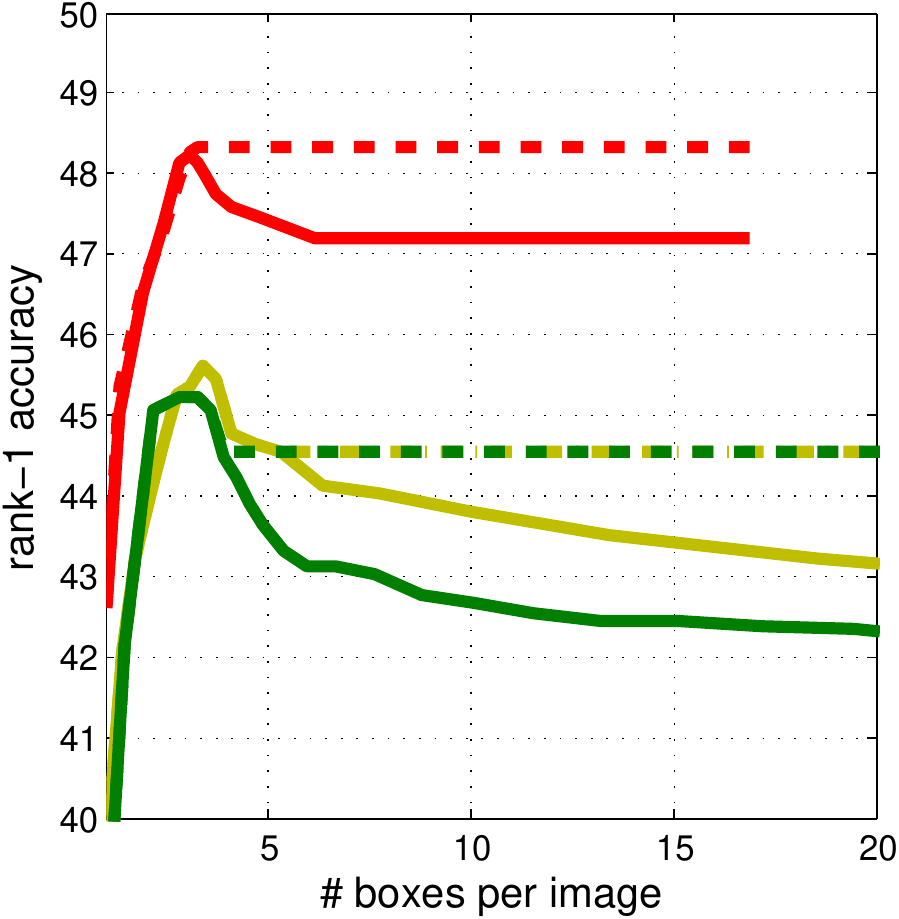}}
\hspace{0.in}
 \subfigure[rank-20, CWS]{\label{fig:rank1_CWS}%
\includegraphics[width=1.05in]{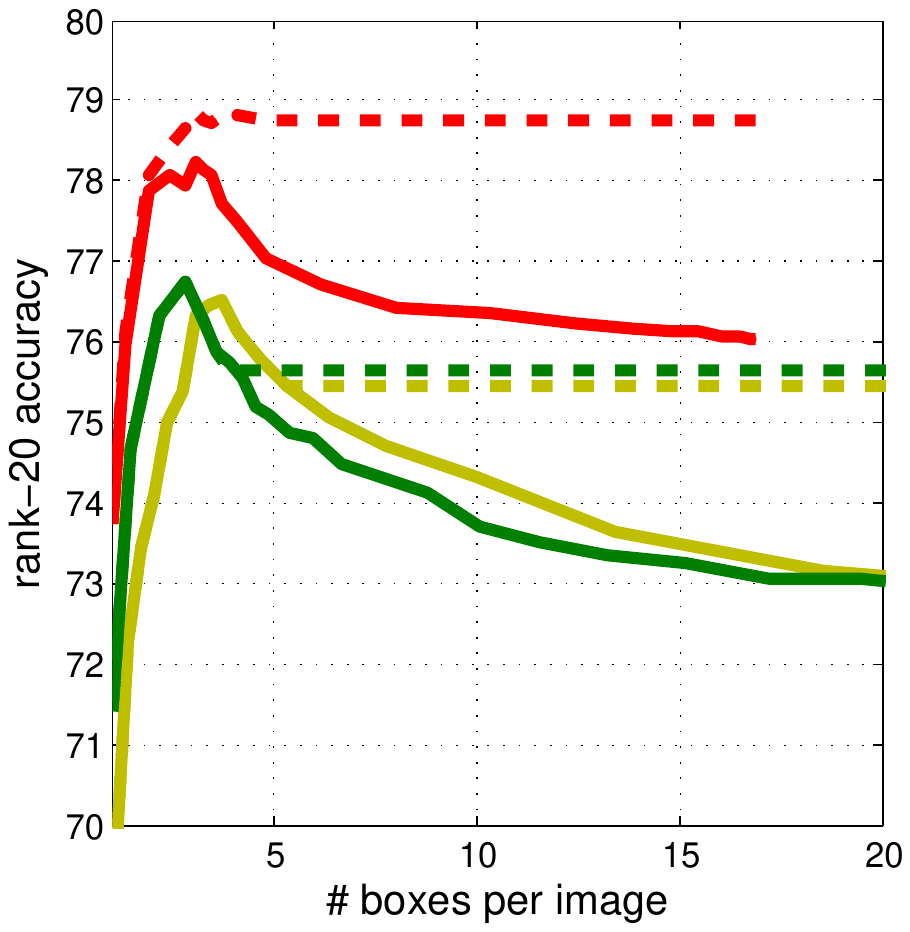}}
\hspace{0.in}
\subfigure[mAP, CWS]{\label{fig:mAP_CWS}%
\includegraphics[width=1.05in]{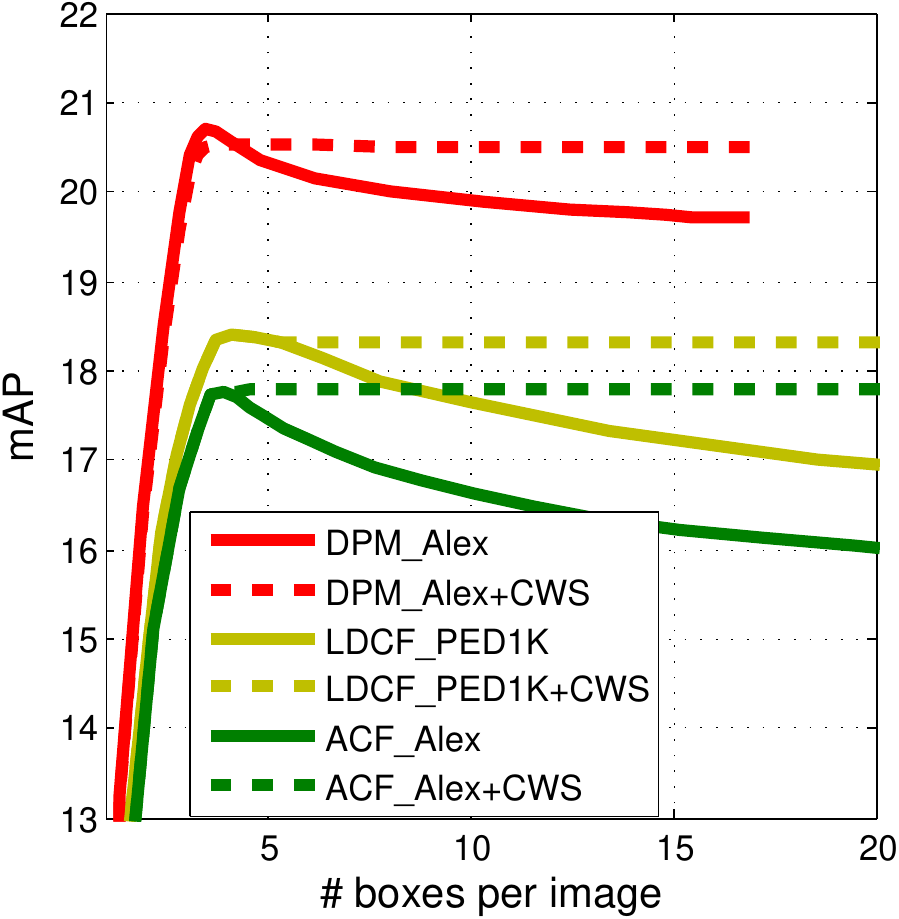}}
\caption{Effectiveness of the proposed Confidence Weighted Similarity (CWS) on the PRW dataset. We test three detectors and the IDE$_{det}$ descriptor fine-tuned on the pedestrian-background detection model. We find that CWS reduces the impact of distractors when the number of detected bounding boxes increases.}
\vspace{-0.4cm}
\label{fig:CWS}
\end {figure}
Two major conclusions can be drawn from the above experiments. First, we observe that the accuracy of IDE is superior to that of hand-crafted descriptors (in accordance with \cite{zheng2016survey}), and is further improved in combination with state-of-the-art metric learning schemes. Second, it is noticeable from Fig. \ref{fig:IDE} that IDE$_{det}$ yields considerably higher re-ID accuracy than IDE$_{imgnet}$. Specifically, when using the DPM detector trained on INRIA dataset and considering 3 detection boxes per image, IDE$_{det}$ results in $+4.52\%$ and $+9.17\%$ improvement in mAP and rank-1 accuracy, respectively. Very similar improvements can be observed for the other 4 detectors and using 5 detection boxes per image. This indicates that when more background and pedestrian samples are ``seen'', the re-ID feature is more robust against outliers. This illustrates that the proposed cascaded fine-tuning method is effective in improving the discriminative ability of the learned embeddings. In fact, a promising direction is to utilize more background and pedestrian samples without ID that are cheaper to collect in order to pre-train the IDE model. Experiment of the two IDE variants provides one feasible solution of how detection aids re-ID.

{\bf Effectiveness of Confidence Weighted Similarity (CWS)} We test the CWS proposed in Section \ref{sec:improvements} on the PRW dataset with three detectors and the IDE$_{det}$ descriptor. The results are shown in Fig.~\ref{fig:CWS}. The key observation is that CWS is effective in preventing re-ID accuracy from dropping as the number of detections per image increase. As discussed before, more distractors are present when the database get larger and CWS addresses the problem by suppressing the scores of false positive results. In Table \ref{table:reid_summary}, the best results on the PRW dataset are achieved when CWS is used, which illustrates the effectiveness of the proposed similarity. We will extend CWS to include metric learning representations in the future work.

\begin{figure}[t]
  \centering
  \includegraphics[width=3.2in]{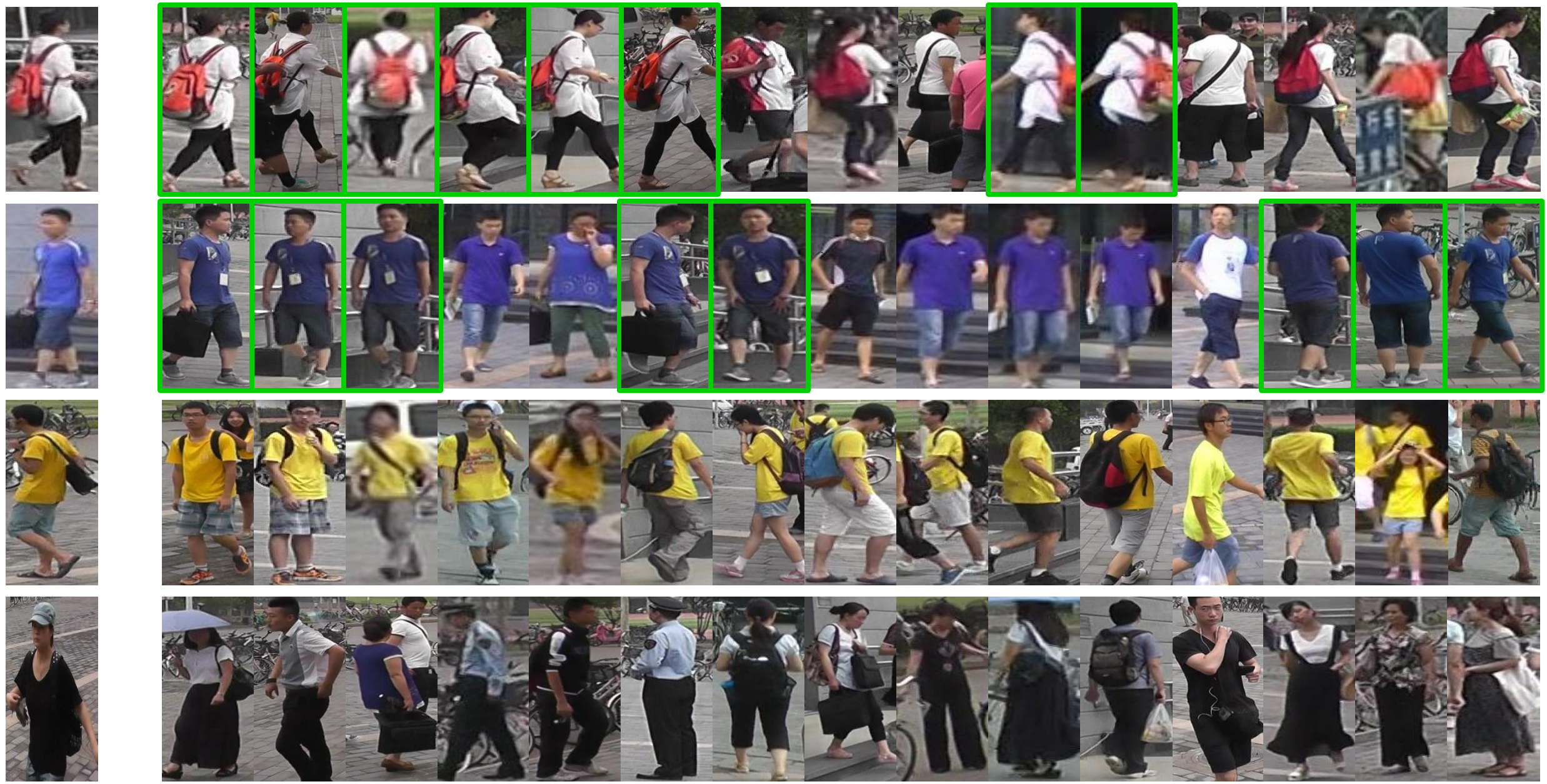}\\
  \caption{Sample re-ID results on the proposed PRW dataset with the DPM\_Alex detector and the proposed IDE descriptor. Rows 1 and 2 are success cases, while Rows 3 and 4 are failure cases due to similar clothing and truncation, respectively. With truncated queries, partial re-ID methods \cite{zheng2015partial} might be especially important.}
 \label{fig:sample_results}
\end{figure}

Figure \ref{fig:sample_results} presents some sample re-ID results. For the failure case in row 3, the reason is that too many pedestrians are wearing similar clothes. For row 4, the query is cropped by the camera, leading to compromised pedestrian matching.

\section{Conclusions and Future Work}
\label{sec:conclusion}

We have presented a novel large-scale dataset, baselines and metrics for end-to-end person re-ID in the wild. The proposed PRW dataset has a number of features that are not present in previous re-ID datasets, allowing the first systematic study of how the interplay of pedestrian detection and person re-ID affects the overall performance. Besides benchmarking several state-of-the-art methods in the fields of pedestrian detection and person re-ID, this paper also proposes two effective methods to improve the re-ID accuracy, namely, ID-discriminative Embedding and Confidence Weighted Similarity. For IDE, we find that fine-tuning an R-CNN model can be a better initialization point for IDE training. Further, our extensive experiments serve as a guide to selecting the best detectors and detection criteria for the specific application of person re-ID.

Our work also enables multiple directions for future research. First, it is critical to design effective bounding box regression schemes to improve person matching accuracy.  Second, given the baseline method proposed in this paper to incorporate detection confidence into similarity scores, more sophisticated re-weighting schemes can be devised. This direction could not have been enabled without a dataset that jointly considers detection and re-ID. In fact, re-ranking methods \cite{zhong2017re,bai2017scalable,zheng2015query} will be critical for scalable re-ID. Third, while it is expensive to label IDs, annotation of pedestrian boxes without IDs is  easier and large amounts of pedestrian data already exist. According to the IDE results reported in this paper, it will be of great value to utilize such weakly-labeled data to improve re-ID performance. Finally, effective partial re-ID algorithms \cite{zheng2015partial} can be important for end-to-end systems on the PRW dataset (Fig. \ref{fig:sample_results}).

{\small
\bibliographystyle{ieee}
\bibliography{egbib}
}

\end{document}